%% file: main.tex
\documentclass[runningheads]{llncs}
\usepackage{graphicx}
\usepackage{amsmath,amssymb} 
\usepackage{booktabs}
\usepackage{color}
\usepackage{xspace}
\usepackage{cite}
\usepackage{adjustbox}
\usepackage{esint} 
\usepackage{algorithm}
\usepackage{placeins}
\usepackage[]{algpseudocode}
\usepackage{siunitx}
\usepackage{subcaption}
\usepackage{hyperref}
\hypersetup{
  colorlinks=true,
  citecolor=[rgb]{0.0, 0.7, 0.0},
  linkcolor=[rgb]{0.8, 0.0, 0.0}
}
\usepackage[font=small,skip=0pt]{caption}

\newcommand{\eat}[1]{}

\begin{document}
\pagestyle{headings}
\mainmatter
\def\ECCVSubNumber{6637}  


\title{SimPose: Effectively Learning DensePose and Surface Normals of People from Simulated Data}
\titlerunning{SimPose}

\author{Tyler~Zhu \and Per~Karlsson \and Christoph~Bregler}
\authorrunning{T. Zhu et al.}

\institute{Google Research\\
\email{\{tylerzhu, perk, bregler\}@google.com}}

\maketitle

\input{abstract}
\input{intro}
\input{related_work}
\input{sp_dataset}
\input{method}

\input{evaluation}
\input{conclusion}

%
%
\bibliographystyle{splncs04}
\bibliography{egbib}
\end{document}

%% file: abstract.tex
\begin{abstract}

With a proliferation of generic domain-adaptation approaches, we report a simple yet effective technique for learning difficult per-pixel 2.5D and 3D regression representations of articulated people. We obtained strong sim-to-real domain generalization for the 2.5D DensePose estimation task and the 3D human surface normal estimation task. On the multi-person DensePose MSCOCO benchmark, our approach outperforms the state-of-the-art methods which are trained on real images that are densely labelled. This is an important result since obtaining human manifold's intrinsic $uv$ coordinates on real images is time consuming and prone to labeling noise. Additionally, we present our model's 3D surface normal predictions on the MSCOCO dataset that lacks any real 3D surface normal labels. The key to our approach is to mitigate the ``Inter-domain Covariate Shift" with a carefully selected training batch from a mixture of domain samples, a deep batch-normalized residual network, and a modified multi-task learning objective. Our approach is complementary to existing domain-adaptation techniques and can be applied to other dense per-pixel pose estimation problems.

\eat{For the first time, we show a model can achieve state-of-the-art performance on the challenging in-the-wild multi-person DensePose COCO benchmark using only simulated dense pose annotations. 
We have created a simulated multi-person dense human pose dataset and we present an effective method to learn a fine-grained dense human pose estimation model. It generalizes to in-the-wild images using only computer graphics simulated annotations of human body parts and body part UV together with real annotations of 17 sparse 2D body joints and person instance segmentation. The proposed method mixes samples across real and simulated domains in the same training batch and adopts a novel multi-task partial loss.  This encourages the learned filters on the core simulated dense pose task to be simultaneously reusable by other verification tasks (2D body joints and person instance segmentation) defined in both real and simulated domains.

Quantitatively we evaluate our training strategy and the resulting model on the DensePose COCO benchmark under the official Geodesic Point Similarity (GPS) metrics. Our model achieved an average precision of 57.3 on the DensePose COCO minival split. This is better than the state-of-the-art average precision of 55.8 reported by the cascaded DensePose-RCNN which was trained in a fully supervised manner using real DensePose annotations plus COCO body joints and person instance segmentation. Finally, we qualitatively visualize our model's prediction on in-the-wild images.}

\keywords{Person pose estimation, simulated data, dense pose estimation, 3D surface normal, multi-task objective, sim-real mixture}

\end{abstract} 

%% file: intro.tex
\newlength{\introFigHeight}
\setlength{\introFigHeight}{2.1cm}
\newlength{\introFigHeightA}
\setlength{\introFigHeightA}{2.23cm}

\section{Introduction}

Robustly estimating multi-person human body pose and shape remains an important research problem. Humans, especially well trained artists, can reconstruct densely three dimensional multi-person sculptures with high accuracy given a RGB reference image. Only recently\cite{pifuSHNMKL19,SMPL-X:2019,Xiang_2019_CVPR,alldieck19cvpr,livecap2019,Guler2018DensePose,NIPS2018_8061,varol18_bodynet} it was demonstrated that this challenging task can also be done using convolutional neural networks with modest in-the-wild accuracy. Improving the accuracy may lead to new applications such as telepresence in virtual reality, large-scale sports video analysis, and 3D MoCap on consumer devices. The challenges presented in natural images of people include mixed clothing and apparel, individual fat-to-muscle ratio, diverse skeletal structure, high-DoF body articulation, soft-tissue deformation, and ubiquitous human-object human-human occlusion.  \\

We explore the possibility of training convolutional neural networks to learn the 2.5D and 3D human  representations directly from renderings of animated dressed 3D human figures (non-statistical), instead of designing an increasingly more sophisticated labeling software or building yet another statistical 3D human shape model with simplifications (e.g., ignoring hair or shoes). However it's well-known that ``deep neural networks easily fit random labels"\cite{Zhang17}. As our simulated 3D human figures are far from being photorealistic, it's easy for the model to overfit the simulated dataset without generalizing to natural images. Therefore the challenge lies in carefully designing a practical mechanism that allows training the difficult 2.5D and 3D human representations on the simulated people domain and generalize well to the in-the-wild domain where there are more details present, e.g., clothing variance (e.g. shoes), occlusion, diverse human skin tone, and hair.

\begin{figure*}
  \begin{center}
  \centering
  \adjustbox{height=\introFigHeight}
      {\includegraphics[trim={0cm 0cm 0cm 0cm},clip]{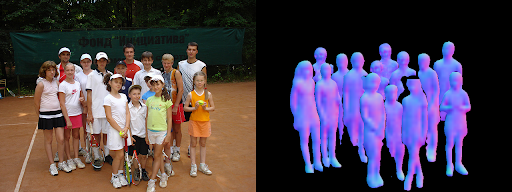}}
  \adjustbox{height=\introFigHeight}
      {\includegraphics[trim={0cm 0cm 0cm 0cm},clip]{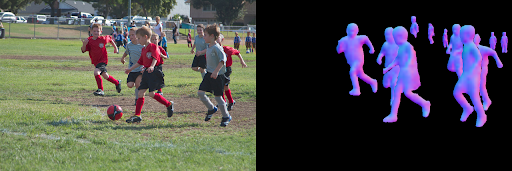}}
  \adjustbox{height=\introFigHeightA}
      {\includegraphics[trim={0cm 0cm 0cm 0cm},clip]{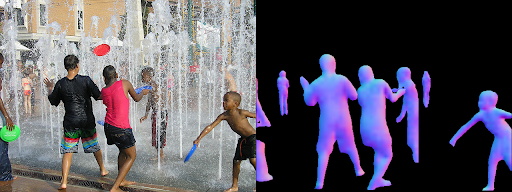}}
  \adjustbox{height=\introFigHeightA}
      {\includegraphics[trim={0cm 0cm 0cm 0cm},clip]{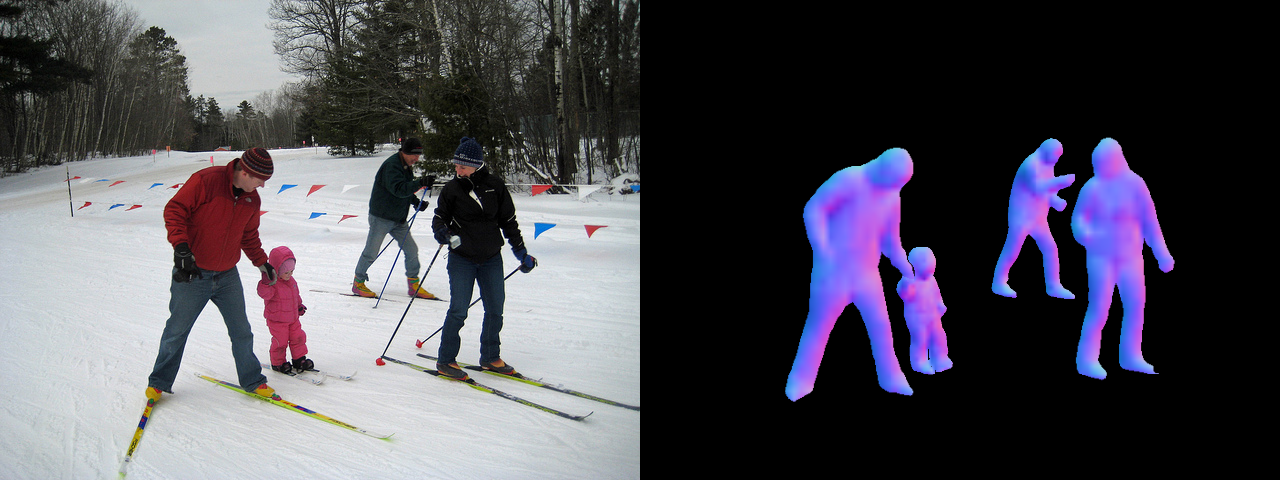}}
  \end{center}
  \caption{Visualization of SimPose's 3D surface normal predictions on MSCOCO dataset.}
  \label{fig:visualization}
\end{figure*}

Deep Convolutional Neural Networks (ConvNets) have been successfully applied to human pose estimation ~\cite{deeppose,jainiclr2014,tompsonnips2014,cmu_mscoco}, person instance segmentation~\cite{He2017MaskR,papandreou2018personlab,chen2016deeplab}, and dense pose estimation~\cite{Guler2018DensePose} (the representation consists of human mesh's continuous intrinsic $uv$  coordinates). To reach the best task accuracy, they all require collecting real human annotated labels for the corresponding task\cite{Guler2018DensePose,keypointchallenge}. As the research frontier for human pose and shape estimation expands into predicting new 2.5D and 3D human representations, it becomes self-evident that labelling large-scale precise groundtruth 2.5D and 3D annotations becomes significantly more time-consuming or perhaps even impossible. \\

Several generic domain-adaptation approaches have been proposed to leverage computer graphics generated precise annotations for in-the-wild tasks: 1) Using Generative Adversarial Network (GAN) framework \cite{Bousmalis2017UnsupervisedPD} to jointly train a generator styled prediction network and a discriminator to classify whether the prediction is made on real or simulated domain. 2) Using Domain Randomization \cite{tobin2017domain} to enhance the input image and thus expand the diversity and horizon of the input space of the model. Our approach is complementary to them. For instance, the GRU-domain-classifier \cite{DomainAdversarial} or the output space discriminator \cite{OutputSpace} can be added as additional auxiliary tasks in our multi-task objective, and more uniform, sometimes corrupting, domain randomization can be added on top of our mild data augmentation. \\

In contrast, we create a new multi-person simulated dataset with accurate dense human pose annotations, and present a simple yet effective technique which mitigates the ``Inter-domain Covariate Shift". The closest prior works \cite{varol17_surreal,varol18_bodynet} to ours in the human pose estimation field utilize a statistical 3D human shape model SMPL \cite{SMPL:2015} for training (building SMPL requires more than 1K 3D scans and geometric simplifications e.g., ignoring hair, shoes). Our approach uses just 17 non-statistic 3D human figures and outperforms SOTA models on DensePose MSCOCO benchmark without additional few-shot real domain fine tuning. \\

In the following sections of the paper, we discuss related works and describe our SimPose approach in more detail: our simulated dataset, system implementation, experiment setup, the accuracy comparison with SOTA models on DensePose MSCOCO benchmark, and 3D human surface normal results. Our main contributions in the paper are the following:
\begin{itemize}
  \item Create \textbf{simulated multi-person datasets} and present a \textbf{simple yet effective batch mixture training strategy with multi-task objective} to learn 2.5D dense pose $uv$ and 3D surface normal estimation model without using any real dense labels. 
  \item Evaluate our proposed approach on the DensePose MSCOCO benchmark \cite{Guler2018DensePose}. We attained \textbf{favourable results using only simulated human $\mathbf{uv}$ labels, better than state-of-the-art models} \cite{Guler2018DensePose,Neverova_2019_CVPR} which are trained using real DensePose labels. We also present our model's 3D surface normal predictions on MSCOCO dataset which lacks any 3D surface normal labels.
  \item Show our approach trained using \textbf{only 17 non-statistical} 3D human figures can obtain better accuracy on the above benchmark than using SMPL\cite{SMPL:2015} (a statistical 3D human shape model built with more than 1K 3D scans). This is a big reduction in number of 3D human scans required. 
\end{itemize} 

%% file: related_work.tex
\section{Related Work}

Human pose estimation and its extension has been a popular research topic over the years. 
Early approaches study the structure of the human skeleton with a Pictorial Structure model (PS model)~\cite{Fischler73}.
The method is later improved by combining it with a probabilistic graphics model~\cite{andriluka2009pictorial,yang11cvpr}.
With the advent of deep learning, the methods are being simplified and the outputs are becoming richer and finer, as we discuss below. 

\textbf{ConvNet Human Pose Estimation}
Toshev and Szegedy~\cite{deeppose} applies convolutional neural networks on the 2D human pose estimation task by regression to the $(x, y)$ locations of each body joint in a cascaded manner. 
Tompson et al.~\cite{tompsonnips2014} proposes a fully convolutional structure by replacing the training target from two scalar (x/y) coordinates to a 2D probabilistic map (or heatmap). 
The robust heatmap based pose representation became popular and has been extended in many different ways. 
Wei et al.~\cite{wei2016convolutional} trains a multi-stage fully convolutional network where the prediction from previous stages are fed into the next stage.
Newell et al.~\cite{stackedhourglass} proposes an encoder-decoder structured fully convolutional network (called HourglassNet) with long-range skip connection to avoid the dilemma between the output heatmap resolution and the network receptive field.
Papandreou et al.~\cite{papandreou2017towards} augments the heatmap with an offset map for a more accurate pixel location. 
Chen et al.~\cite{chen2017cascaded} proposes a cascaded pyramid structured network with hard keypoint mining.
Recently, Bin et al.~\cite{xiao2018simple} shows that state-of-the-art 2D human pose estimation performance can be achieved with a very simple de-convolutional model. 

\textbf{3D Human Pose \& Shape Estimation Tasks}
The progress on 2D human pose estimation motivates the community to move forward to more high-level human understanding tasks and take advantage of the 2D pose results.
Bottom-up multi-person pose estimation~\cite{cmu_mscoco,newell2017associative,papandreou2018personlab} simultaneously estimates the pose of all persons in the image, by learning the instance-agnostic keypoints and a grouping feature. 
3D human pose estimation~\cite{zhou2016sparseness,zhou2017towards,martinez2017simple} learns the additional depth dimension for the 2D skeleton, by learning a 3D pose dictionary~\cite{zhou2016sparseness} or adding a 3D bone length constraint~\cite{zhou2017towards}.
Human parsing~\cite{fang2018weakly,liang2016semantic} segments a human foreground mask into more fine-grained parts. DensePose~\cite{Guler2018DensePose, Neverova_2019_CVPR} estimation aims to align each human pixel to a canonical 3D mesh. Kanazawa et al.~\cite{hmrKanazawa17} predicts plausible SMPL coefficients using both 2D reprojection loss and 3D adversarial loss. Kolotouros et al.~\cite{kolotouros2019spin} improves it by incorporating SMPL fitting during training. Xu et al.~\cite{DenseRac} shows that using pre-trained DensePose-RCNN's predictions can further improve 3D pose and shape estimation accuracy.

\textbf{Simulated Synthetic Datasets}
When collecting training data for supervised learning is not feasible, e.g., for dense per-pixel annotation or 3D annotation, simulated training data can be a more useful solution. 
Chen et al.~\cite{chen2016synthesizing} synthesizes 3D human pose data by deforming a parametric human model (SCAPE~\cite{anguelov2005scape}) followed by adding texture and background. 
Similarly, the SURREAL dataset~\cite{varol17_surreal} applies the CMU mocap motion data to the SMPL human model~\cite{loper2015smpl} on SUN~\cite{xiao2010sun} background, with 3D pose, human parsing, depth, and optical flow annotations. Zheng et al.~\cite{DeepHuman} utilizes a Kinect sensor and DoubleFusion algorithm to capture posed people and synthesizes training images which contain one person per image.
Compared to the previous simulated human datasets, our datasets contain multi-person images and two 3D human model sources (non-statistical and statistical) for comparison. 

\textbf{Domain Adaptation}
As the simulated synthetic data is from a different distribution compared with the real testing data, domain adaptation is often applied to improve the performance. 
A popular approach is domain adversarial training~\cite{tzeng2017adversarial}, which trains a domain discriminator and encourages the network to fool this domain discriminator. This results in domain invariant features. 
Domain randomization~\cite{tobin2017domain} generates an aggressively diverse training set which varies all factors instead of the training label, and causes the network to learn the inherited information and ignore the distracting randomness. 
Other ideas incorporate task specific constraints. Contrained CNN~\cite{pathak2015constrained} iteratively optimizes a label prior for weakly supervised semantic segmentation in a new domain, and Zhou et al.~\cite{zhou2017towards} enforce bone length constraint for 3D human pose estimation in the wild. In this paper, we propose a simpler alternative that generalizes well. 

%% file: sp_dataset.tex
\section{Simulated Multi-person Dense Pose Dataset}

\newlength{\myheight}
\setlength{\myheight}{3cm}
\newlength{\myheightB}
\setlength{\myheightB}{2cm}
\newlength{\myheightO}
\setlength{\myheightO}{1.8cm}

\begin{figure*}
  \begin{center}
  \centering
  \adjustbox{height=\myheight}
      {\includegraphics[trim={0cm 0cm 0cm 0cm},clip]{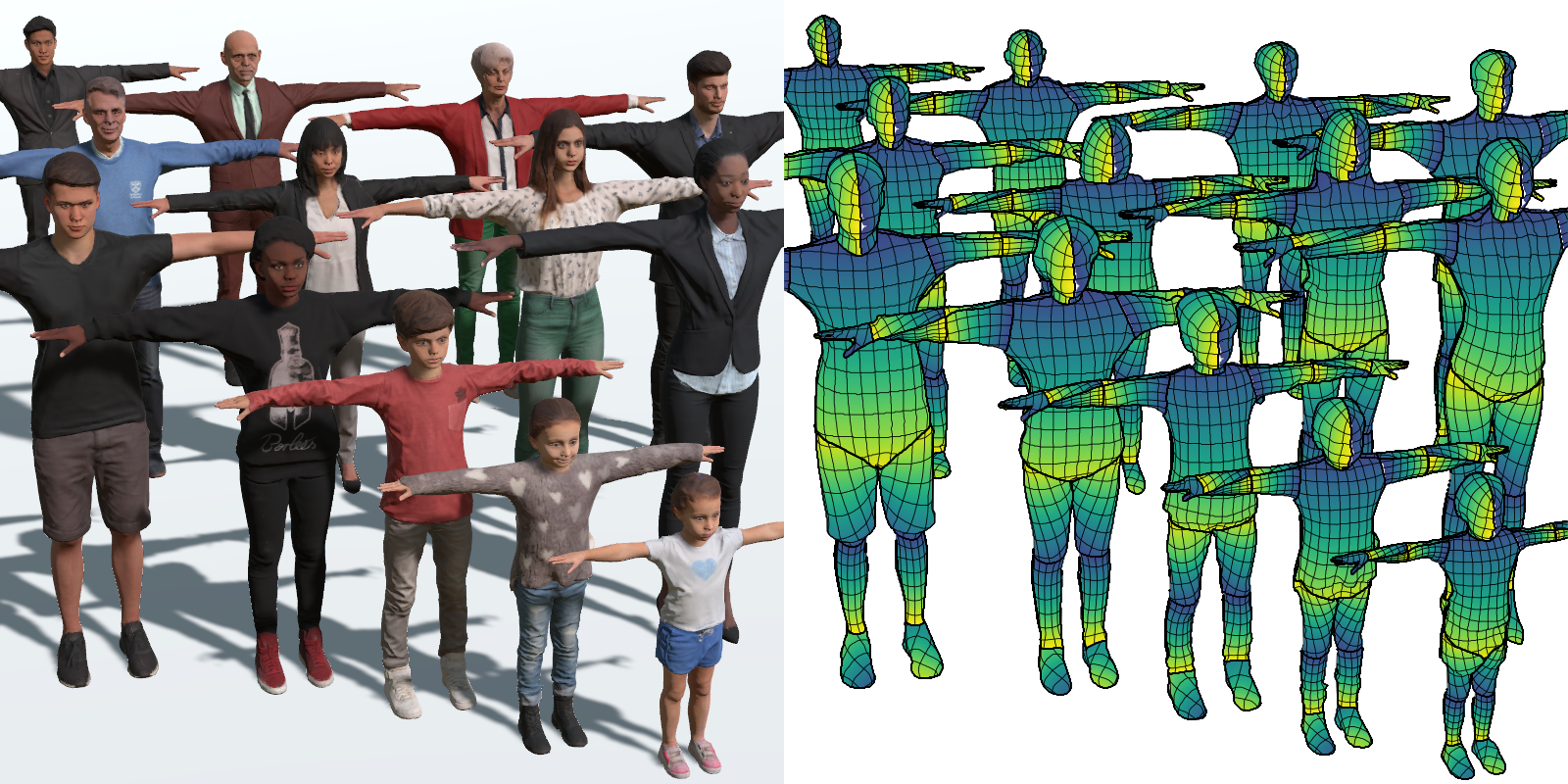}}
  \adjustbox{height=\myheight}
      {\includegraphics[trim={0cm 0cm 0cm 0cm},clip]{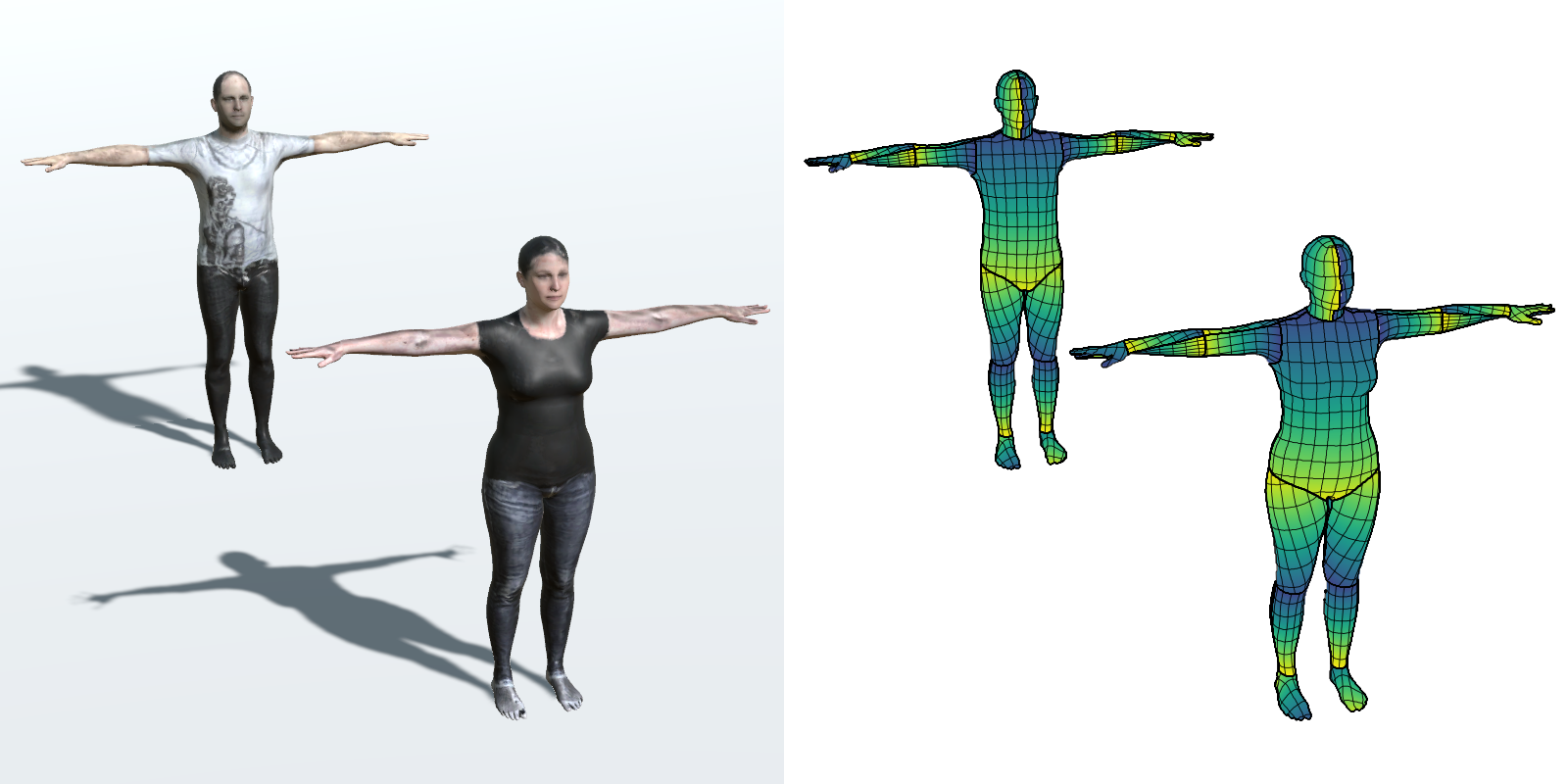}}
  \end{center}
  \caption{Visualization of the 3D human model sources used to create our two simulated people datasets. Renderpeople 3D figures to the left and SMPL with SURREAL textures to the right. For SMPL, we only visualize its two base shapes here, but we sample shapes from its continuous shape space in our experiments. The human manifold's intrinsic $uv$ coordinates labels are visualized as isocontour lines where the background is the dense value of $v$. }
  \label{fig:3d_models}
\end{figure*}

Collecting dense continuous labels on real images is not only expensive and time-consuming but also hard to label accurately. By creating a simulated synthetic dataset, we are able to get consistent per pixel 2.5D and 3D human pose and shape labels (body part segmentation, body part intrinsic $uv$ coordinates, and 3D surface normals), in comparison to~\cite{Guler2018DensePose} where the annotations are not per-pixel labelled. This section describes how we prepare our human 3D models with the non trivial task of generating correct $uv$ mapping.

\subsection{Human 3D Models}
\label{human_3d_models}
We create and compare two simulated datasets with two different sources of human 3D models, one consisting of only 17 rigged high resolution 3D scans of humans acquired from Renderpeople~\cite{renderpeople} and one using the statistical SMPL body model ~\cite{loper2015smpl} (built from 1785 scans) together with SURREAL surface textures~\cite{varol17_surreal}. Other differences between the sources are the visual quality of the meshes and textures. The Renderpeople models include clothing and hair while the SMPL model does not. The Renderpeople textures are more realistic while the SURREAL textures may have artifacts. 3D reference points are attached to the skeleton of each rigged human model. We align these points with the MSCOCO 2D human keypoint definition for rendering simulated 2D keypoint.

\textbf{Render People}
We ensure that all our human 3D models have the same $uv$ and body part segmentation mapping as the references in DensePose labels~\cite{Guler2018DensePose} so that our approach can be evaluated on the MSCOCO DensePose benchmark. This is not easy to achieve as there are no available tools out there. We propose Algorithm \ref{alg:MatchUV}, which takes about 40 min per Renderpeople model (total 17 3D human model) for a person with previous experience in 3D modeling software. 

\begin{algorithm}
\caption{Transfer DensePose UV to Renderpeople meshes}\label{alg:MatchUV}
\begin{algorithmic}
\For{each bodypart submesh $bp_{ref}$ in reference mesh }
\State Mark a subset of the boundary vertices in ${bp}_{ref}$ as landmarks $L_{ref}$. 
\State Store UV coordinates $UV_{ref}$ for each vertex in $L_{ref}$.
\EndFor

\For{each Renderpeople mesh $rp$}
\For{each bodypart submesh $bp_{ref}$ in reference mesh}
  \State Cut a submesh $bp_{rp}$ from $rp$ that is similar to $bp_{ref}$.
  \State Mark the same landmark vertices $L_{rp}$ as in $L_{ref}$.
  \State Copy UV coordinates from $UV_{ref}$ to $UV_{rp}$.
  \State Linearly interpolate UV coordinates along the boundary vertices in $bp_{rp}$.
  \State Mark boundary vertices in $bp_{rp}$ as static.
  \State Unwrap UV on the inner non-static vertices, similar to~\cite{unwrap_uv}.
\EndFor
\EndFor

\end{algorithmic}
\end{algorithm}

\textbf{SMPL} It is trivial to populate the DensePose UV coordinates from~\cite{Guler2018DensePose} since they also use the SMPL model as the reference mesh.

\begin{figure*}
  \begin{center}
  \centering
  \adjustbox{height=\myheightB}
      {\includegraphics[trim={0cm 0cm 0cm 0cm},clip]{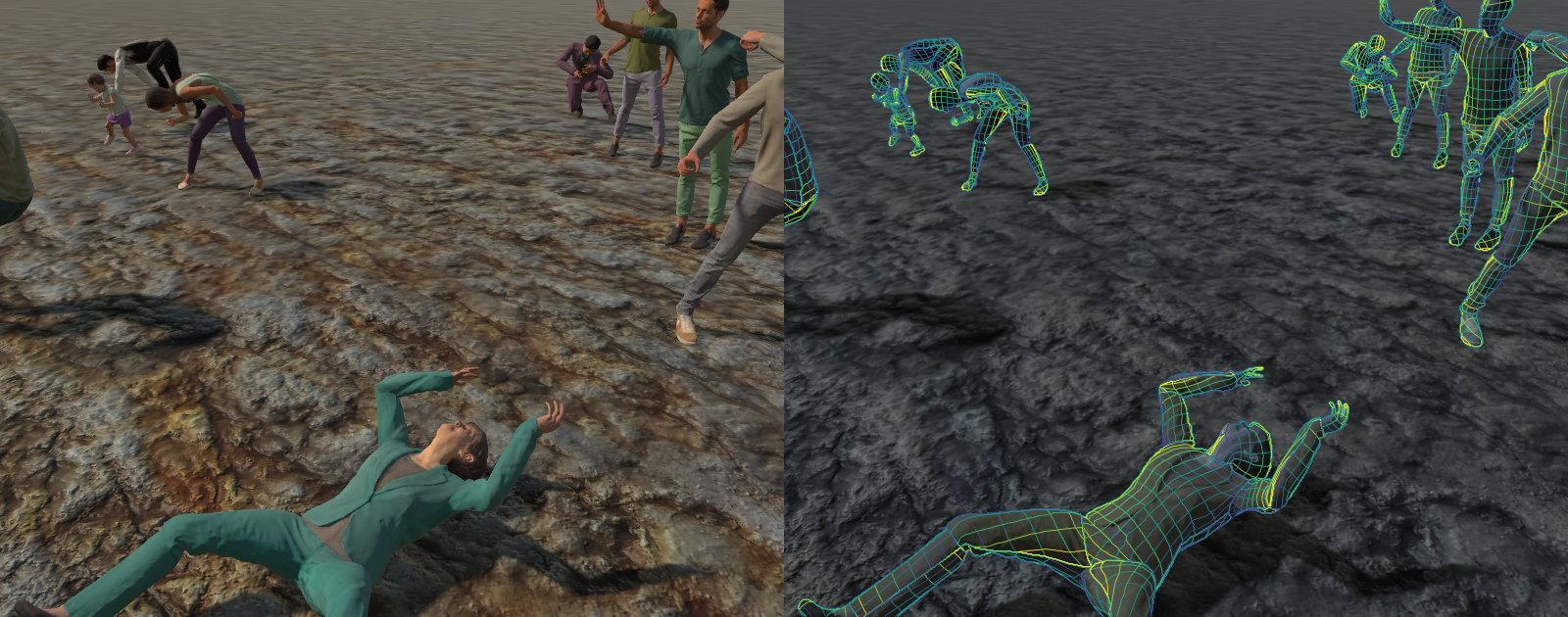}}
  \adjustbox{height=\myheightB}
      {\includegraphics[trim={0cm 0cm 0cm 0cm},clip]{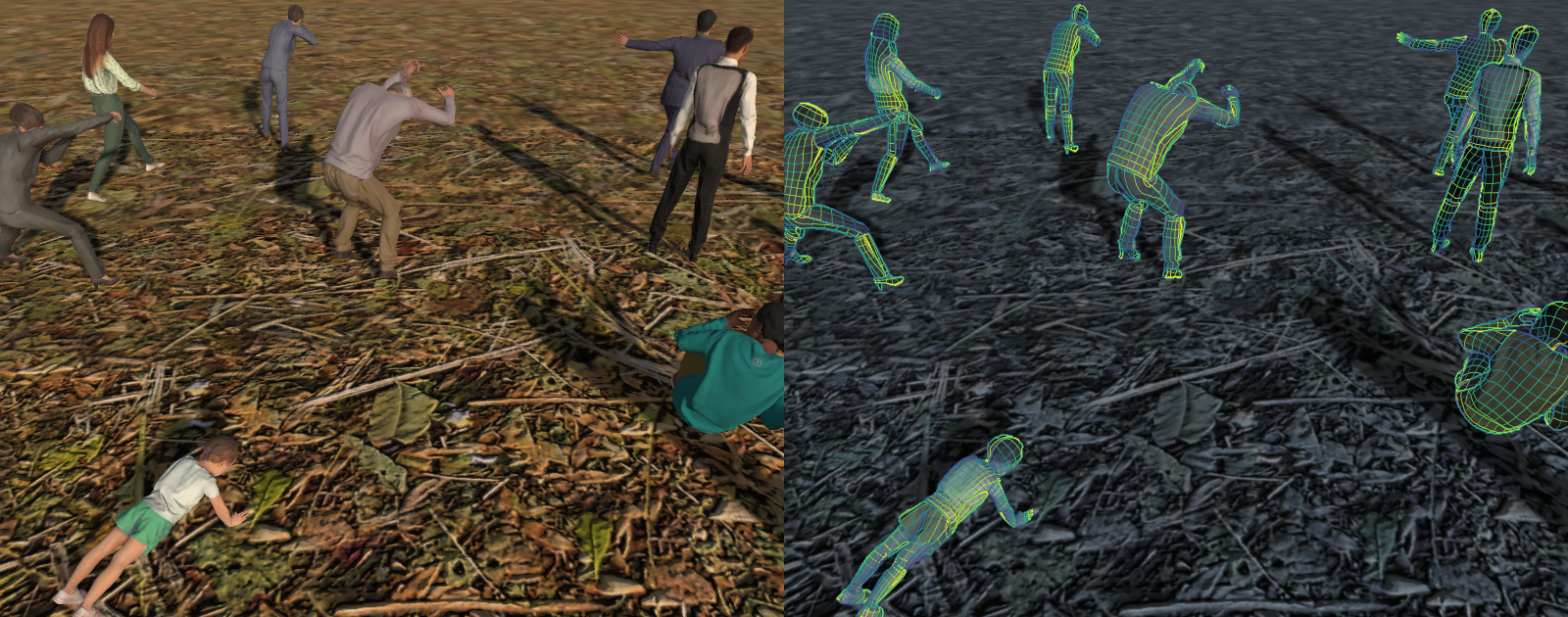}}
  \centering
  \adjustbox{height=\myheightB}
      {\includegraphics[trim={0cm 0cm 0cm 0cm},clip]{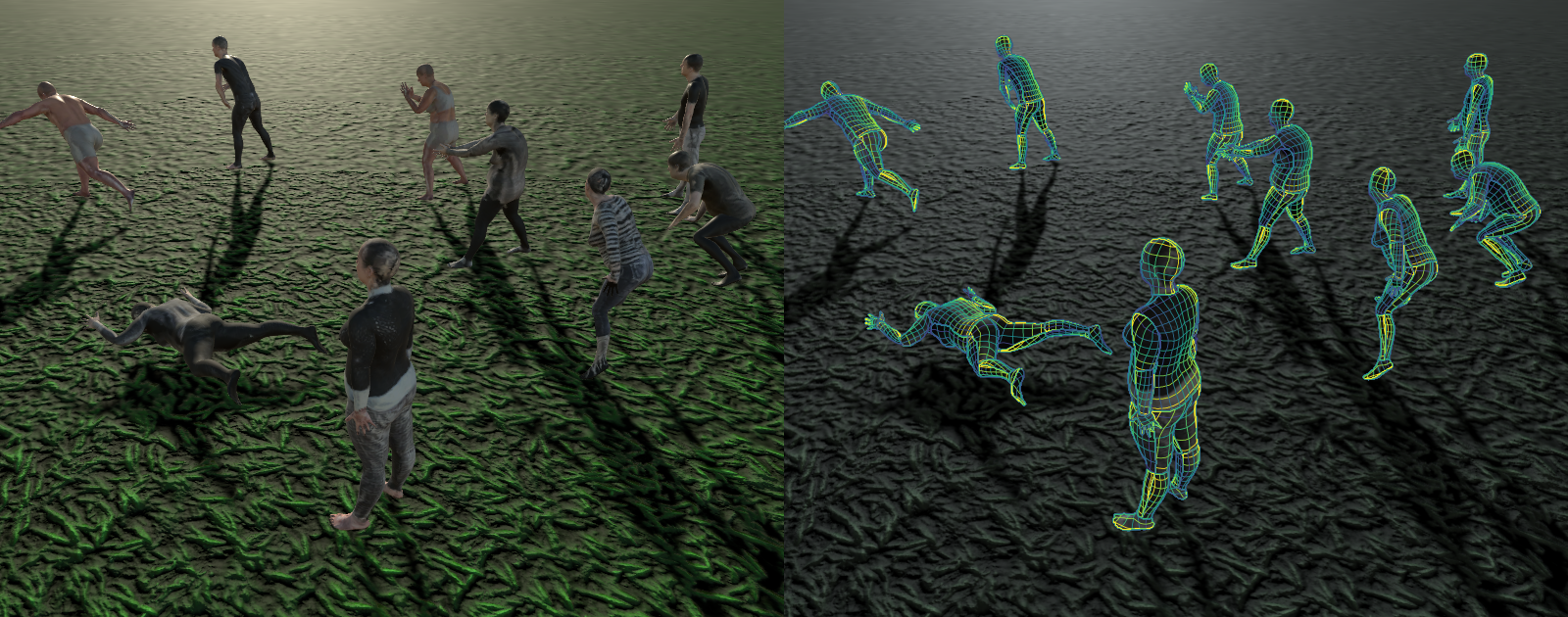}}
  \adjustbox{height=\myheightB}
      {\includegraphics[trim={0cm 0cm 0cm 0cm},clip]{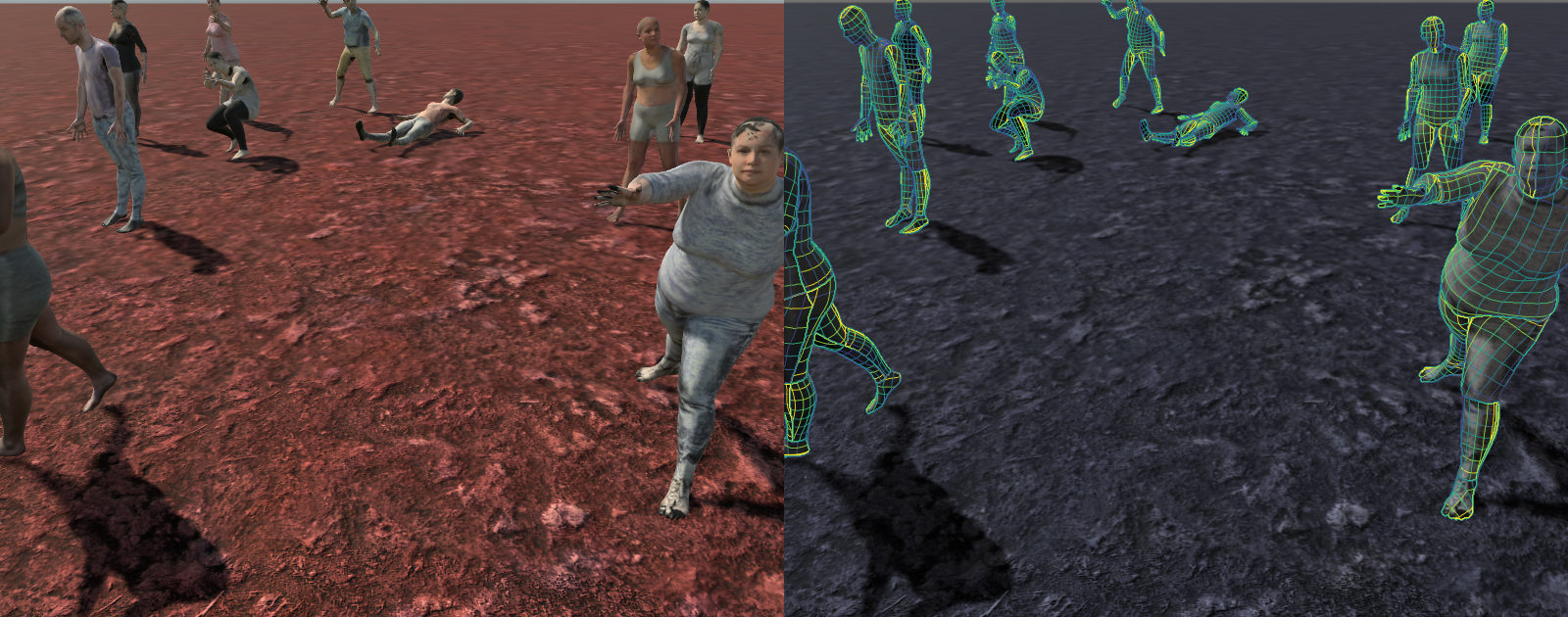}} 
  \end{center}
  \caption{Examples from our simulated datasets. Top row shows the Renderpeople dataset and bottom row shows the SMPL dataset. Only the $uv$ label is shown.}
  \label{fig:dataset_examples}
\end{figure*}

To make it easier to train and compare our approach on the datasets separately, we make sure that each image only has humans from one of the sources. Figure~\ref{fig:dataset_examples} shows example images from both sources. In total, we generated 100,000 images for each source.

\begin{figure*}
  \begin{center}
  \centering
  
  \adjustbox{height=\myheightO}
      {\includegraphics[trim={0cm 0cm 0cm 0cm},clip]{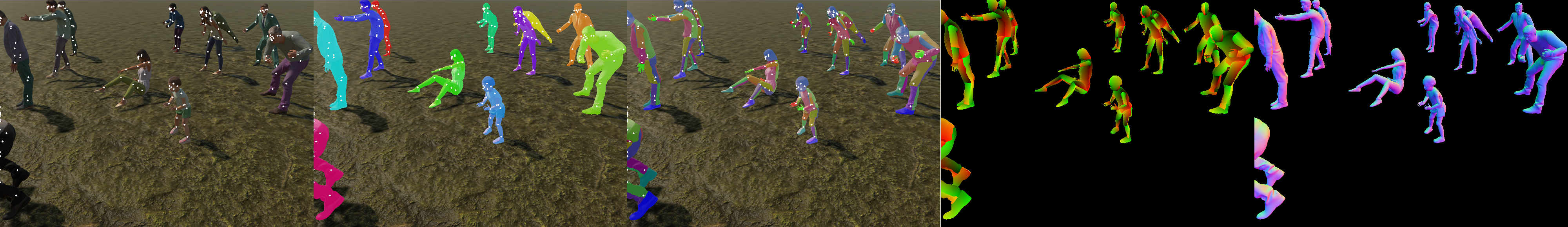}}
  \end{center}
  \caption{Examples from our Renderpeople simulated datasets. From left to right: 2D keypoints, instance segmentation, body parts, $uv$, and 3D surface normal.}
  \label{fig:dataset_examples_full}
\end{figure*}

%% file: method.tex
\section{Our Approach}

Given a two-stage multi-person pose estimation meta-architecture, our approach, as illustrated by Figure \ref{fig:pose_model}, anchors the domain adaptation of the 2nd-stage model's new 2.5D and 3D human pose \& shape estimation tasks on a set of easier 2D tasks. \eat{and utilizes multi-modal whitening layers to permit domain-agonostic features that are robust to cross-domain covariate shift, if it's the optimal to do.} Inspired by the recent works of multi-task \& meta-learing \cite{MAML,MetaReg,GradientSurgery}, a modified domain-label-based multi-task objective trains a deep normalized network end-to-end on a batch mixture of sim-real samples. We show that within a deep network, batch normalization, together with convolutions, can permit whitening and aligning bi-modal distributions via Batch Mixture Normalization. From an empirical perspective, our approach: \textbf{i}. generalizes from only 17 non-statistical 3D human figures, which we show in Table~\ref{table:denseposecoco_results_minival}, and hence suffices as a functional replacement of using a statistical human shape model built from $>$1k 3D scans; \textbf{ii}. is able to directly domain-adapt continuous intrinsic $uv$ coordinates and 3D surface normals. 
\eat{\textbf{c}. achieves domain-generalization without belt-and-wistle (e.g., hard-craft causality in the architecture~\cite{varol18_bodynet} or extra fewshot real-label fine-tuning\cite{varol17_surreal}).} 

\input{meta_architecture}

\input{pose_estimation_tasks}

\input{partial_loss}

%% file: meta_architecture.tex
\subsection{Human Pose Estimation Meta-architectures}
\label{person_detection_model}
Among two widely used meta-architectures for multi-person human pose \& shape estimation tasks, one is the bottom-up meta-architecture and the other is the two-stage meta-architecture. In the bottom-up meta-architecture, a neural network is applied fully convolutionally to predict instance-agnostic body parts for all people, as well as additional auxiliary representations for grouping atomic parts into individual person instances (e.g., part affinity field\cite{cmu_mscoco} and mid-range offsets\cite{papandreou2018personlab}). The two-stage meta-architecture relies on a separate detection network or an RPN (region proposal network) whose outputs, ROIs (Region of Interest) of the input image or of the intermediate feature maps, are then cropped, resized, and fed into a 2nd stage pose estimation network. 

We adopt the two-stage meta-architecture. The crop-and-resize operation is used as spatial attention of the second-stage network to restrict the divergence between real and simulation distributions within the ROIs. More specifically, our two-stage meta-architecture is the following: the 1st stage is a Faster RCNN person detector with ResNet101\cite{He2016ResNets} backbone (following \cite{papandreou2017towards}, we make the same surgery to the original Faster RCNN including using dilated convolutions and training only on MSCOCO\cite{lin2014microsoft} person bounding box alone); The 2nd stage pose estimation network is shown in Figure \ref{fig:pose_model}. Before applying the corresponding tasks' losses, the outputs from the 2nd stage model are upsampled to be of the same size as the input image crops using the differentiable bilinear interpolation. \eat{Similar to \cite{papandreou2017towards}, we add additional supervision at intermediate layer 50 of ResNet101 to accelerate training and we modify the ResNet101 backbone to use atrous convolution with the output stride set to 8.}

%% file: pose_estimation_tasks.tex
\subsection{Human Pose Estimation Tasks}
\label{section:all_tasks}
We adopt the term ``task" that is used unanimously in the meta-learning, multi-task learning, and reinforcement learning literature. We cluster the human pose \& shape estimation tasks into two categories: \textbf{i}. the 2D tasks and \textbf{ii}. the 2.5D/3D tasks. The 2D tasks are well-studied and stabilized, and they are: 2D body keypoint localization and 2D person instance segmentation. The 2.5D/3D tasks are new, and we select the following two tasks: human mesh intrinsic $uv$ coordinates regression and 3D surface normal vector regression. \eat{The reason being the piece-wise surface geodesic $uv$ parameterization (a Hilbert Space Isomorphisms), if also ``flattenable"\cite{Sharp:2018:VSC}, together with the 3D surface normal vector field $\mathbf{n}(u,v)$ can be used to approximately reconstruct the surface, up to a rigid transformation.} Below we describe each task and the task formulation in details.

\newlength{\heightFigure}
\setlength{\heightFigure}{5cm}

\begin{figure*}
    \begin{center}
    \centering
    \adjustbox{height=\heightFigure} {
    \includegraphics{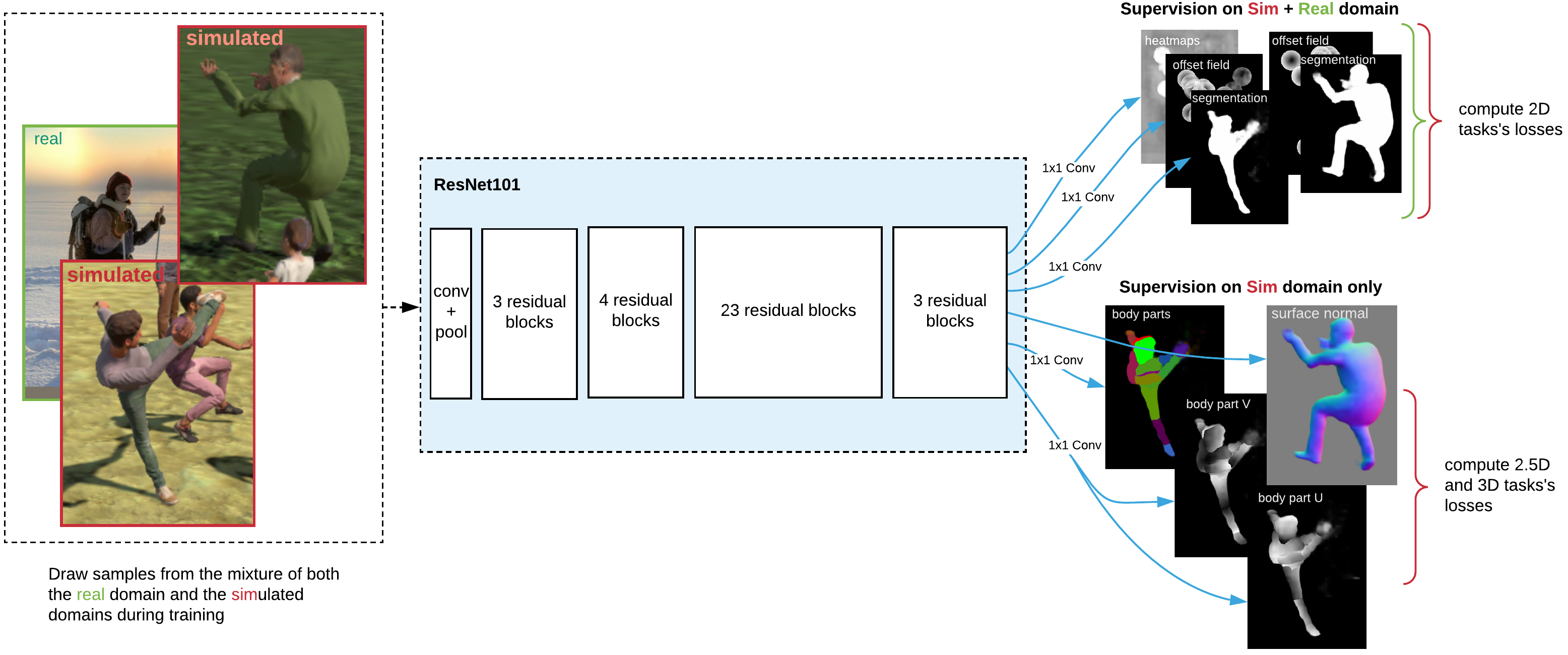}
    }
    \end{center}
    \caption{The ResNet101 network takes image crops batch mixture from both real and simulated domains as input and generates predictions for both 2.5D 3D tasks and 2D tasks which consist of sparse 2D body keypoints and instance segmentation prediction tasks. We compute the 2D tasks' losses for both the real and the simulated domains and compute the 2.5D and 3D tasks' losses only for the simulated domain.}
    \label{fig:pose_model}
\end{figure*}

\textbf{2D Human Keypoints Estimation Task} 

The sparse 2D human keypoint estimation task requires precisely localizing the 2D subpixel positions of a set of keypoints on the human body (e.g., 17 keypoints defined by MSCOCO: ``left shoulder", ``right elbow", etc). We follow \cite{papandreou2017towards} and let the 2nd stage network predict heatmaps and offset fields. We use Hough-voting\cite{papandreou2017towards} to get more concentrated score maps. The final 2D coordinates of each keypoint are extracted via the spatial argmax operation on the score maps. The heatmap predictions are supervised using per-pixel per-keypoint sigmoid cross entropy loss and the offset field is supervised using per-pixel per-keypoint Huber loss. 

\textbf{2D Person Instance Segmentation Estimation Task}

The task of instance segmentation requires classifying each pixel into foreground pixels that belong to the input crop's center person, and background pixels that don't. We follow \cite{chen2016deeplab} and let the 2nd stage network predict per-pixel probability of whether it belongs to the center person or not, and use standard per-pixel sigmoid cross entropy loss to supervise the prediction.

\textbf{2.5D Body Part Intrinsic $UV$ Coordinates Regression Task}

For the 2.5D $uv$ coordinates regression task, following the work of \cite{Guler2018DensePose}, we predict $K$ body part segmentation masks and $2K$ body $uv$ maps in the DensePose convention (e.g., ``right face", ``front torso", ``upper right leg", and etc). We use per-pixel per-part sigmoid cross entropy loss to supervise the body part segmentation and used per-pixel per-part smooth $l_{1}$ loss to supervise the $uv$ predictions. We only backprop the $l_{1}$ loss through the groundtruth body part segmentation region. For the $uv$ maps we subtract 0.5 from the groundtruth value to center the range of regression.

\textbf{3D Person Surface Normal Regression Task}

For the 3D person surface normal task, the 2nd stage network directly regresses per-pixel 3D coordinates of the unit surface normal vectors. During post-training inference, we $l_{2}$ renormalize the 2nd-stage networks's per-pixel 3D surface normal predictions: $\mathbf{\hat{n}} = \frac{\mathbf{n}}{||\mathbf{n}||}$. We use per-pixel smooth $l_{1}$ loss to supervise the surface normal predictions. We only backprop the loss $L$ through the groundtruth person instance segmentation region $S$ using a discrete variance of: 
\begin{equation}
  \frac{\partial L(\theta)}{\partial \theta} = \frac{\partial}{\partial \theta} \sum_{i}^m \iint_{S_i} ||\mathbf{n} - \mathbf{f}_{\theta}(I_i)|| \,dx\,dy
  \label{eq:1}
\end{equation}
where $\theta$ is all the trainable parameters of the network, $m$ is the batch size, $I_i$ is the $i$-th image in the batch, $(x,y)$ are the coordinates in the image space, $\mathbf{n} = (n_x, n_y, n_z)$ is the groundtruth 3D surface normal vector at $(x,y)$, $\mathbf{f}_\theta(\cdot)$ is the network predicted surface normal vector at $(x,y)$, $S_i$ is the support of the groundtruth person segmentation in image $I_i$, and $||\cdot||$ is the smooth $l_1$ norm.

%% file: partial_loss.tex
\subsection{Multi-task Learning and Batch Mixture Normalization }

Learning to estimate people's 2.5D intrinsic $uv$ coordinates and 3D surface normals from renderings of 17 simulated people alone, and generalizing to natural image domain of people, requires a deep neural network to learn domain-invariant representations of the human body. We utilize a shared backbone trained end-to-end to achieve this. The network needs to learn domain-invariant representations of a 3D human body at its higher layers that are robust to the inevitably diverged representations at its lower layers. We refer to this underlying issue as the Inter-domain Covariate Shifts.

Batch Normalization~\cite{BatchNorm} was designed to reduce the``Internal Covariate Shift"\cite{UnderstandBN,BatchNBias} for supervised image classification tasks. It assumes similar activation statistics within the batch and whitens the input activation distribution by its batch mean $\mu$ and batch variance $\sigma$. When training a shared backbone using a batch mixture of images sampled from the two domains of a) simulated people and b) natural images of people, the two input activation distributions almost certainly don't share the same mean $\mu$ and variance $\sigma$. Below we show one non-trivial realization of multi-modal whitening using: 1. Batch Normalization, 2. Fuzzy AND ($\land$) OR ($\vee$) implemented as convolution followed by activation, and 3. implicitly learnt domain classifier within the network. We call it Batch Mixture Normalization (BMN), Algorithm \ref{alg:BMN}. During training, we believe the deep batch-normalized shared backbone can whiten bi-modal activations from the sim-real batch mixture using a fuzzy variance of BMN, when it's the optimal thing to do:

\begin{algorithm}
\caption{Batch Mixture Normalization}\label{alg:BMN}
\textbf{Require:} $s^k(\cdot)$: a k-th layer featuremap (useful only on sim images) \\
\textbf{Require:} $r^k(\cdot)$: a k-th layer featuremap (useful only on real images) \\
\textbf{Require:} $z^k(\cdot)$: a k-th layer featuremap implicitly learnt as domain classifier \\
\textbf{Require:} Fuzzy AND ($\land$) OR ($\vee$) implemented as convolution followed by activation \\

\begin{algorithmic}
\State Sample a batch of $m$ image $\{I_i\}_i^m \sim \mathcal{D}_{sim} \cap \mathcal{D}_{real}$
\State Evaluate k layers to get: $\{z\}_i^m, \{s\}_i^m, \{r\}_i^m = z^+(\{I_i\}_i^m), s^+(\{I_i\}_i^m), r^+(\{I_i\}_i^m)$
\State Mask domain outliers (the k+1 layer): $[...\{s\},\{r\}, ...]= [..., \framebox{$\{z \land s\}, \{\bar{z} \land r\}$} ...]$  
\State Batch Normalize (the k+2 layer): $[..., \{s\}, \{r\}, ...] = BatchNorm([..., \{s\},\{r\}, ...])$ 
\State Bring into the same channel (the k+3 layer): $[..., \{y\}, ...] = [..., \framebox{$\{s \vee r\}$}, ...]$ \\
\State \Comment{For succeeding layers ($\geqslant$ k + 4), feature $\{y\}_i^m$ is the whitened and aligned version of $s^k(\{I_i\}_i^m)$ and $r^k(\{I_i\}_i^m)$}.

\end{algorithmic}
\end{algorithm}

 We incorporate domain labels in the multi-task objective to cope with the constraint that 2.5D human intrinsic $uv$ and 3D human body surface normal labels are only available on the simulated domain. Intuitively each task demands that the network learns more about the human body structure and shape in a complementary way. Due to mixing and normalizing, the network's forward and backward passes depend on the composition of the sim-real batch mixture.
 
\begin{equation}
  L(\{I_i\}, \theta) = \sum_{l=1}^m L^{2D}(f_{\theta}(\{I_i\})) + \sum\limits_{\substack{j=1 \\ I_j \in \mathcal{D}_{sim}}}^m [L^{2.5D}(f_{\theta}(\{I_i\}) + L^{3D}(f_{\theta}(\{I_i\}))] 
  \label{eq:3}
\end{equation}
where $L^{2D}$ is the loss term for the 2D keypoint estimation task and the 2D instance segmentation task, $L^{2.5D}$ is the loss term for the human intrinsic $uv$ coordinates regression task, and $L^{3D}$ is the loss term of the 3D human body surface normal regression task. $f_{\theta}(\cdot)$ is the shared human pose and shape backbone parameterized by $\theta$, $m$ is the batch size, and $\{I_i\}_i^m$ is the batch mixture of simulated image crops and real image crops. In practice, we adopt ResNet101 as our backbone and apply a set of $1\times1$ convolutions at the last layer of ResNet101 to predict: keypoint heatmaps, keypoint offset field, person instance segmentation, body part segmentations, body part intrinsic $uv$ coordinates, and 3D human surface normals (see Section \ref{section:all_tasks}). \eat{To cope with the fact that we only have fine-grained dense pose annotations on simulated images but not on real images, we propose to use partial loss to supervise the training examples coming from both real domain and simulation domain on multiple tasks. Regardless of its simple form, we are able to learn domain agnostic filters and generalize the dense pose prediction capability learnt from simulation annotations to in-the-wild real images.}  We apply a per task weight to balance the loss for each task. To avoid exhaustive grid-search and task gradient interference~\cite{GradientSurgery}, we adopt a greedy hyper-parameter search for individual task weights: 1) search for 2D task weights: $w_{heatmap}$, $w_{offsets}$, $w_{segment}$. 2) fix the 2D task weights and sequentially search for the weights for $uv$ and normal tasks: $w_{parts}$, $w_{uv}$, and $w_{normal}$. We used the following weights for each task: $w_{heatmap} = 4.0$, $w_{offsets} = 1.0$, $w_{segment} = 2.0$, $w_{parts} = 0.5$, $w_{uv} = 0.25$, and $w_{normal} = 1.00$. During training we apply an additional gradient multiplier (10x) to the $1\times1$ convolution weights used to predict the heatmap, offset field, and person instance segmentation.

Compared with\cite{wu2019detectron2}, our main differences are: 1) we have a new multi-person 3D surface normal prediction (Section \ref{section:all_tasks}) that Detectron2 doesn't have; 2) we obtained competitive DensePose accuracy, learnt from synthetic labels (Section \ref{sec:denseposecoco_onpar_state_of_the_art}); 3) our modified multi-task objective induces a richer label set and gradients for the simulated part (add. 3D surface normal) of the mini-batch, vs a reduced label set and gradients for the real part of the mini-batch; 4) we optimize only the 2nd-stage pose estimator instead of jointly optimizing RPN and multi-heads.

%% file: evaluation.tex
\section{Evaluation}
\label{sec:results}

\renewcommand{\arraystretch}{1.1}
\addtolength{\intextsep}{-15pt}

\begin{table*}
\centering
\caption{UV Performance on DensePose COCO \textbf{minival} split (higher is better).}
\label{table:denseposecoco_results_minival}
\scalebox{0.85}{
\begin{tabular}{@{}lcccccccccc@{}} \hline 
                                                         & AP    & \si{AP_{50}} & \si{AP_{75}} & \si{AP_M} & \si{AP_L} & AR & \si{AR_{50}} & \si{A_{75}} & \si{AR_{M}} & \si{AR_{L}} \\ \hline \hline
\textbf{Prior Works} (use labelled IUV) & \\ \hline 
DP-RCNN-cascade~\cite{Guler2018DensePose}                & 51.6  & 83.9  & 55.2   & 41.9   & 53.4   & 60.4  & 88.9  & 65.3   & 43.3   & 61.6 \\
DP-RCNN-cascade + masks~\cite{Guler2018DensePose}       & 52.8  & 85.5  & 56.1   & 40.3   & 54.6   & 62.0  & 89.7  & 67.0   & 42.4   & 63.3 \\ 
DP-RCNN-cascade + keypoints~\cite{Guler2018DensePose}  & 55.8  & 87.5  & 61.2   & 48.4   & 57.1   & 63.9  & 91.0  & 69.7   & 50.3   & 64.8 \\ 
Slim DP: DP-RCNN (ResNeXt101)~\cite{Neverova_2019_CVPR}              & 55.5  & \textbf{89.1}  & 60.8   & 50.7   & 56.8   & 63.2  & 92.6  & 69.6   & 51.8   & 64.0 \\
Slim DP: DP-RCNN + Hourglass~\cite{Neverova_2019_CVPR}  & \textbf{57.3}  & 88.4  & 63.9   & 57.6   & 58.2   & 65.8  & 92.6  & 73.0   & 59.6   & 66.2 \\ \hline \hline

\textbf{Ours} (use \emph{only} simulated IUV) & \\  \hline 

SimPose (3D model source: Renderpeople)                  & \textbf{57.3} & 88.4  & \textbf{67.3}   & 60.1   & \textbf{59.3}   & \textbf{66.4}  & \textbf{95.1}  & \textbf{77.8}  & 62.4   & \textbf{66.7}  \\
SimPose (3D model source: SMPL)                          & 56.2          & 87.9   & 65.3   & \textbf{61.0}   & 58.0   & 65.2  & 95.1  & 75.2  & \textbf{63.2}  & 65.3  \\ \hline
\end{tabular}}
\end{table*}

\subsection{Experimental Setup}
\label{sec:experimental_setup}
Our SimPose system is implemented in the TensorFlow framework \cite{tensorflow2015-whitepaper}. The 2nd-stage model is trained using the proposed approach. We use 4 P100 GPUs on one single machine and use synchronous stochastic gradient descent optimizer (learning rate is set to $0.005$, momentum value is set to $0.9$, batch size is set to 8 for each GPU). We train the model for $240$K steps. The ResNet101 backbone has been pre-trained for 1M steps on the same sparse 2D $17$ human keypoints and instance segmentation annotations mentioned above (no real MSCOCO DensePose labels are used) from an ImageNet classification checkpoint. The 1st-stage Faster RCNN person detector is trained on the MSCOCO person bounding box labels using asynchronous distributed training with 9 K40 GPUs. Stochastic gradient descent with momentum is used as optimizer (learning rate is set to $0.0003$, momentum value is set to $0.9$, learning rate is decayed by 10x at the $800$K step of the total $1$M training steps). The ResNet101 backbone has been pretrained on the ImageNet classification task.

We use the Unity game engine to generate our simulated people datasets. We render up to 12 humans per image. Each image is 800x800 pixels with 60 degrees field of view and with motion blur disabled. All shaders use the default PBR~\cite{pbr_course} shading in Unity with low metallic and smoothness values. The shadows use a high resolution shadow map with four shadow cascades. The position of each human on the ground plane is random given the constraint that it can not intersect another human. We pose the human models with a random pose from one of 2,000 different Mixamo animations, similar to~\cite{Villegas_2018_CVPR}. \eat{In order to to avoid humans from floating, each mesh is projected down to the ground plane. The SMPL Unity model comes with blend shapes, which can be used to deform the mesh shape while keeping the same topology. This allows us choose between a random shape along with a random SURREAL texture.} The 17 Renderpeople models do not have blend shapes or different textures. Instead we randomly augment the hue~\cite{colorspaces} of the Renderpeople clothing to add variation.

\newlength{\myheightA}
\setlength{\myheightA}{3.5cm}
\newlength{\myheightF}
\setlength{\myheightF}{2.5cm}
\newlength{\myheightC}
\setlength{\myheightC}{2.7cm}
\newlength{\myheightD}
\setlength{\myheightD}{2.6cm}
\newlength{\myheightE}
\setlength{\myheightE}{2.2cm}

\begin{figure*}
  \begin{center}
  \centering
  \adjustbox{height=\myheightA}
      {\includegraphics[trim={0cm 0cm 0.0cm 0cm},clip]{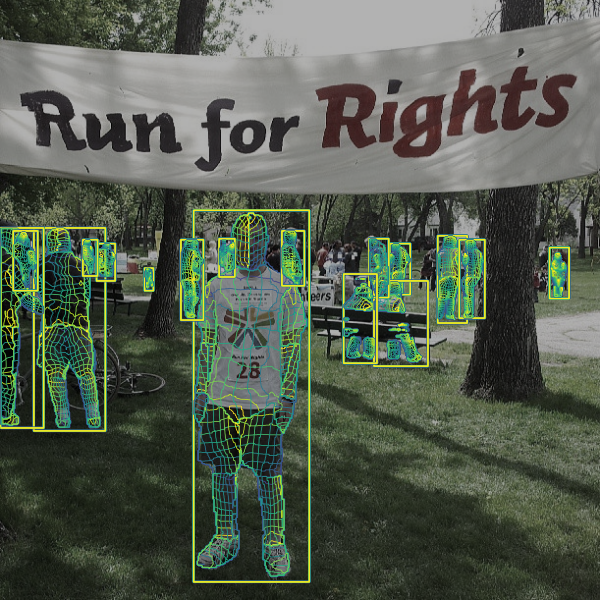}}
  \adjustbox{height=\myheightA}
      {\includegraphics[trim={0cm 0cm 0cm 0cm},clip]{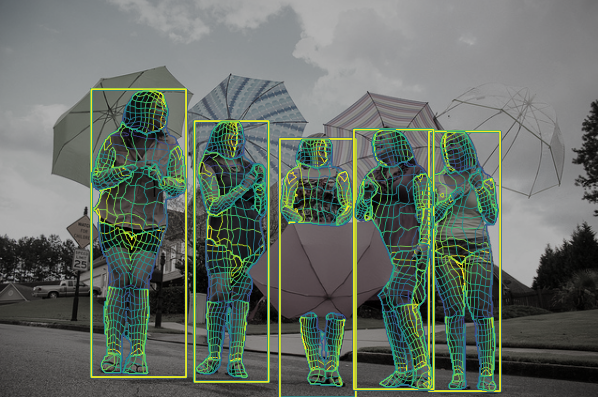}}
  \adjustbox{height=\myheightA}
      {\includegraphics[trim={0cm 0cm 0cm 0cm},clip]{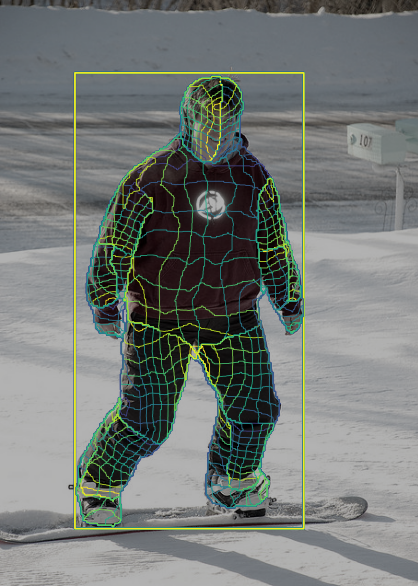}}
  \centering
  \adjustbox{height=\myheightC}
      {\includegraphics[trim={0cm 0cm 0cm 0cm},clip]{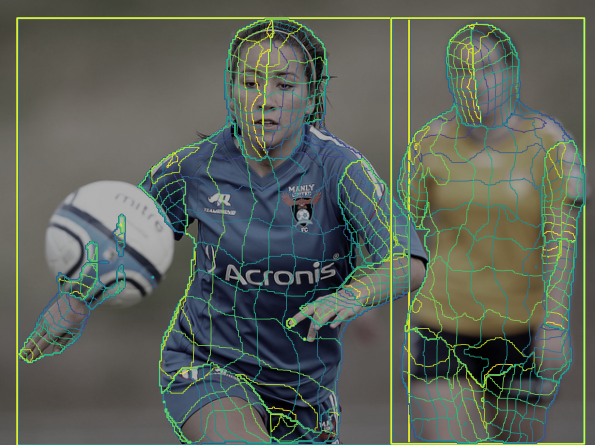}}      
  \adjustbox{height=\myheightC}
      {\includegraphics[trim={0cm 0cm 0cm 0cm},clip]{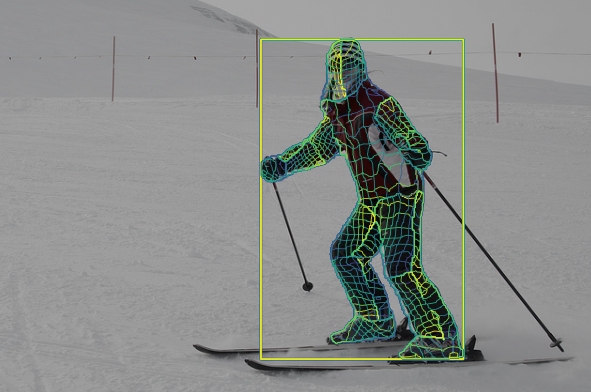}}
  \adjustbox{height=\myheightC}
      {\includegraphics[trim={0cm 0cm 0cm 0cm},clip]{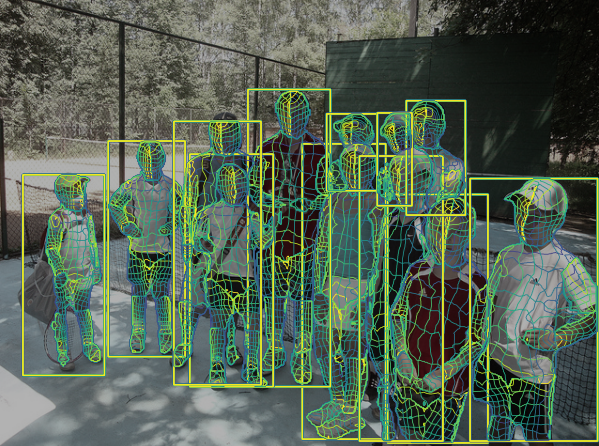}}
  \end{center}
  \caption{SimPose's UV predictions on DensePose MSCOCO minival split.}
  \label{fig:results_array}
\end{figure*}

\subsection{Evaluation of SimPose's UV Prediction}
\label{sec:denseposecoco_onpar_state_of_the_art}
 Table~\ref{table:denseposecoco_results_minival} shows our SimPose approach's accuracy on the DensePose MSCOCO minival split ($1.5K$ images). Despite the fact that our model has not used any real DensePose labels, it achieves 57.3 average precision measured by the GPS (Geodesic Point Similarity) metrics on the challenging multi-person DensePose MSCOCO benchmark, which is better than the DensePose-RCNN\cite{Guler2018DensePose} model's average precision of 55.8, and better than the state-of-the-art Slim DensePose\cite{Neverova_2019_CVPR} on most of the breakdown AP \& AR metrics. Both DensePose-RCNN and Slim DensePose have been trained using real DensePose labels. 
 
 Furthermore, in Table~\ref{table:denseposecoco_results_minival} we also compare the accuracy of our system trained from Renderpeople and SMPL separately. We found using only 17 Renderpeople 3D human models and our proposed approach, the system achieves better performance than using SMPL (a statistical 3D human shape model built from more than 1K 3D human scans). We conduct ablation studies of the simulated data mixing ratio (Table~\ref{table:ablation_mix_ratio}) and the UV task weight (Table~\ref{table:ablation_uv_task_weight}) for the 2nd stage model. In Figure~\ref{fig:results_array}, we visualize our SimPose system's UV predictions on the DensePose MSCOCO minival split.
 
\setlength{\tabcolsep}{12pt}
\begin{table*}
\centering
\caption{Ablation Study of Simulated Data Mixing Ratio. Higher mixing ratio of simulated data is detrimental to model's 2D tasks accuracy on MSCOCO and has diminishing returns for the 2.5D and 3D tasks on the Renderpeople validation set.}
\label{table:ablation_mix_ratio}
\scalebox{0.8}{
\begin{tabular}{@{}lllllll@{}} \hline
        Task        &  Metric  & \multicolumn{5}{c}{Percentage of Sim Data} \\
                              \cmidrule{3-7}
                           &   & 100\% & 75\% & 50\% & 25\% & 0\% \\ \hline  \hline
     Normal & ADD (lower is better) & 19.3\si{\degree} & 19.5\si{\degree} & 19.9\si{\degree} & 21.5\si{\degree} & 74.6\si{\degree}   \\
     UV & L2 (lower is better)  & 0.075   & 0.075  & 0.077  & 0.082  & 0.386 \\
     Segmentation & IOU (higher is better)   & 0.26    & 0.68   & 0.69   & 0.69.  & 0.70 \\
     Keypoint & OKS (higher is better) & 0.32 & 0.74 & 0.76 & 0.75 & 0.77 \\ \hline
\end{tabular}}
\caption{Ablation Study of UV Task Weight. Higher value doesn't affect 2D tasks on COCO and has diminishing returns for UV accuracy on Renderpeople validation set.}
\label{table:ablation_uv_task_weight}
\scalebox{0.8}{
\begin{tabular}{@{}lllllll@{}} \hline
         Task     &  Metric   & \multicolumn{5}{c}{UV Task Weight} \\
                             \cmidrule{3-7}
                     &        & 0.00  & 0.25   & 0.50 & 1.00 & 2.00 \\ \hline  \hline
     UV & L2 (lower is better)  & 0.384 & 0.077  & 0.076  & 0.074  & 0.071 \\
     Segmentation & IOU (higher is better)   & 0.698 & 0.692  & 0.697  & 0.695. & 0.697 \\
     Keypoint & OKS (higher is better)   & 0.761 & 0.756  & 0.760  & 0.760  & 0.756 \\ \hline
\end{tabular}}
\end{table*}

\subsection{Evaluation of SimPose's Surface Normal Prediction}
\label{sec:surface_normal}
 In Figure~\ref{fig:visualization},~\ref{fig:results_normal_array}, and~\ref{fig:results_normal_array_side_by_side}, we qualitatively visualize SimPose's 3D surface normal predictions on a withheld MSCOCO validation set which the model hasn't been trained on. Our model generalizes well to challenging scenarios: crowded multi-person scenes, occlusions, various poses and skin-tones. We evaluate the 3D surface normal predictions on the Renderpeople validation set using ADD (Average Degree Difference). We conduct ablation studies of the simulated data mixing ratio (Table~\ref{table:ablation_mix_ratio}) and the 3D normal task weight (Table~\ref{table:ablation_normal_task_weight}). 
 
\setlength{\tabcolsep}{12pt}
\begin{table*}
\centering
\caption{Ablation Study of 3D Normal Task Weight. Higher value doesn't affect 2D tasks on COCO and has diminishing returns on the Renderpeople validation set.}
\label{table:ablation_normal_task_weight}
\scalebox{0.8}{
\begin{tabular}{@{}lllllll@{}} \hline
        Task     &  Metric    & \multicolumn{5}{c}{3D Normal Task Weight} \\
                              \cmidrule{3-7}
                 &            & 0.00  & 0.25  & 0.50   & 1.00   & 2.00 \\ \hline  \hline
     Normal & ADD (lower is better) & 76.1\si{\degree} & 22.1\si{\degree} & 21.1\si{\degree} & 19.9\si{\degree} & 18.6\si{\degree}  \\
     UV & L2 (lower is better)   & 0.077 & 0.077  & 0.076  & 0.077  & 0.077 \\
     Segmentation & IOU (higher is better)   & 0.699 & 0.697  & 0.700  & 0.696  & 0.698 \\
     Keypoint & OKS (higher is better)  & 0.763 & 0.759  & 0.760  & 0.756  & 0.760 \\ \hline
\end{tabular}}
\end{table*}

\renewcommand{\floatpagefraction}{.9}

\newlength{\myheightAA}
\setlength{\myheightAA}{2.32cm}
\newlength{\myheightCC}
\setlength{\myheightCC}{2.073cm}
\newlength{\myheightDD}
\setlength{\myheightDD}{1.455cm}
\newlength{\myheightEE}
\setlength{\myheightEE}{1.375cm}
\newlength{\myheightFF}
\setlength{\myheightFF}{2.2cm}
\newlength{\myheightGG}
\setlength{\myheightGG}{2.2cm}
\newlength{\HSideBySide}
\setlength{\HSideBySide}{1.35cm}

\begin{figure*}
  \centering
  \adjustbox{height=\myheightAA}
      {\includegraphics[trim={0cm 0cm 0.0cm 0cm},clip]{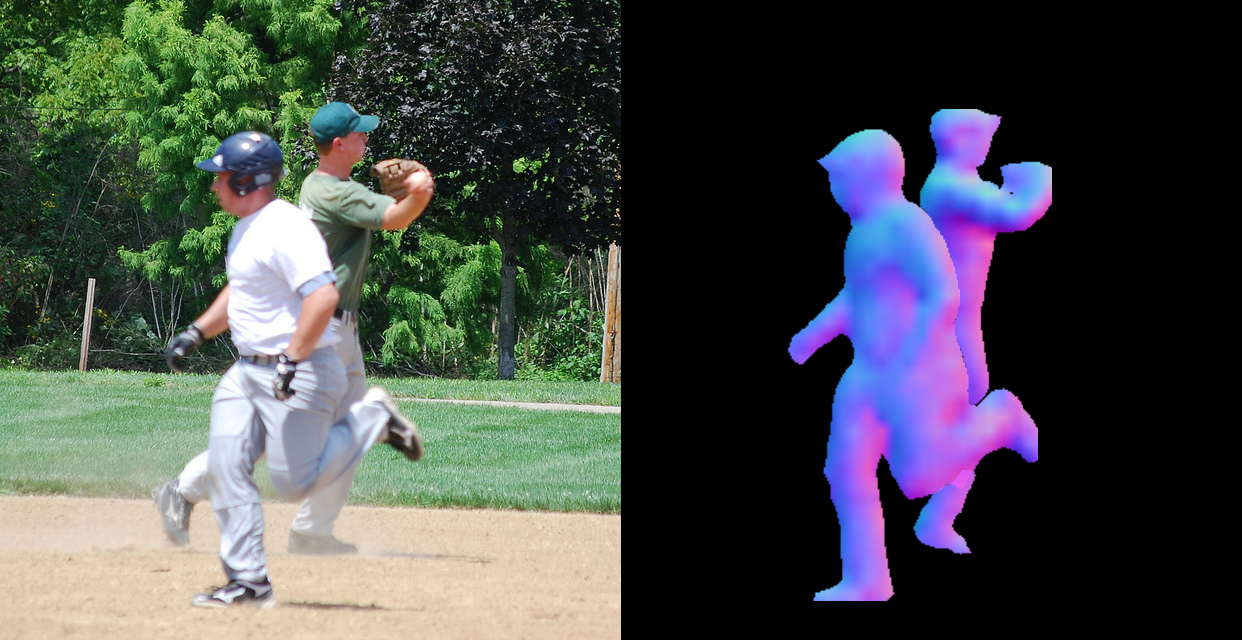}}
  \adjustbox{height=\myheightAA}
      {\includegraphics[trim={0cm 0cm 0cm 0cm},clip]{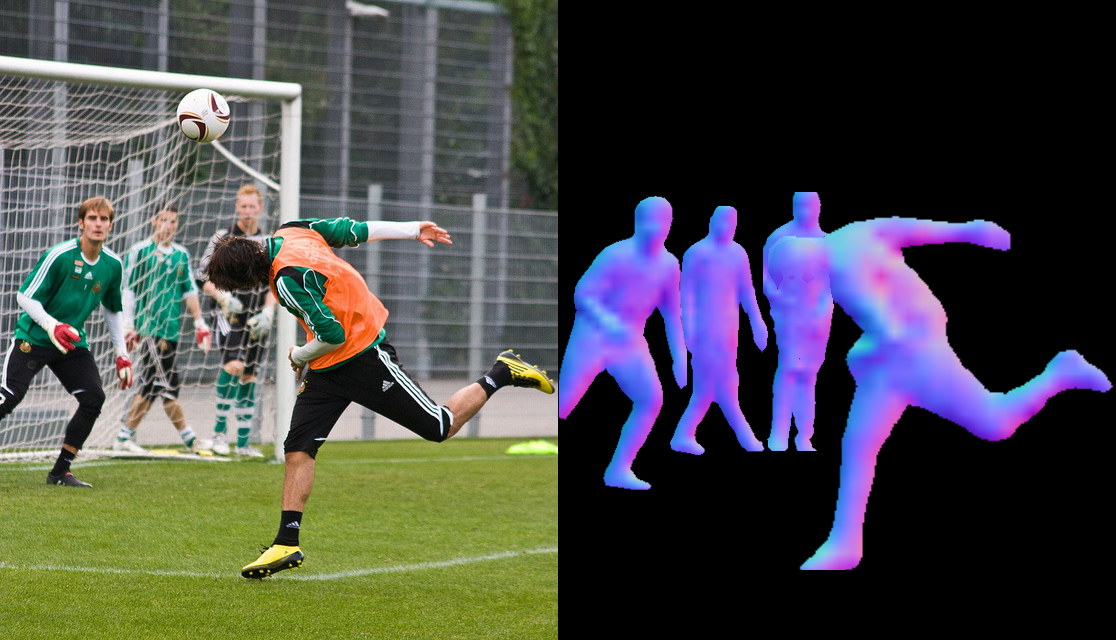}}
  \adjustbox{height=\myheightAA}
      {\includegraphics[trim={0cm 0cm 0cm 0cm},clip]{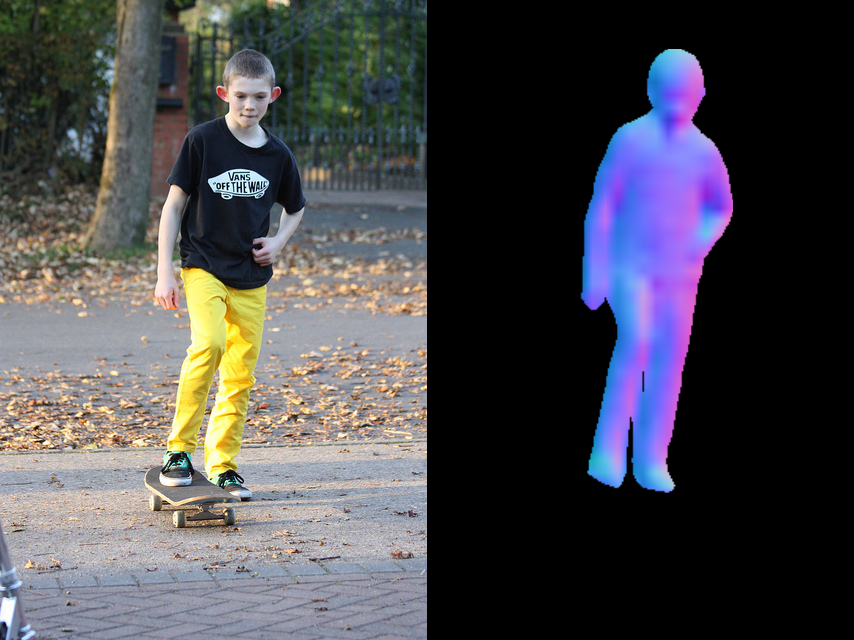}}
  \centering
  \adjustbox{height=\myheightCC}
      {\includegraphics[trim={0cm 0cm 0cm 0cm},clip]{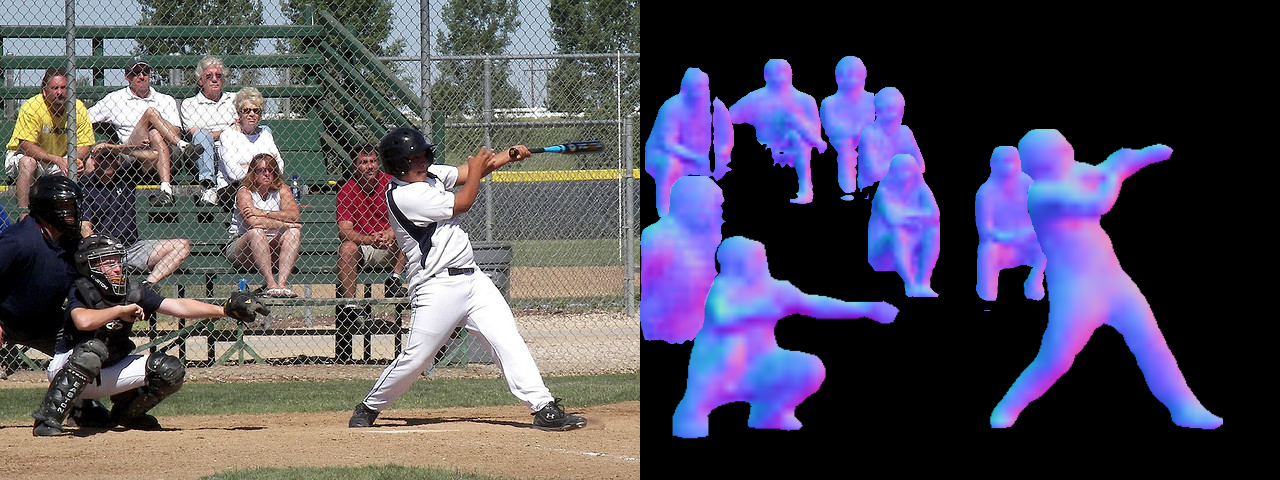}}      
  \adjustbox{height=\myheightCC}
      {\includegraphics[trim={0cm 0cm 0cm 0cm},clip]{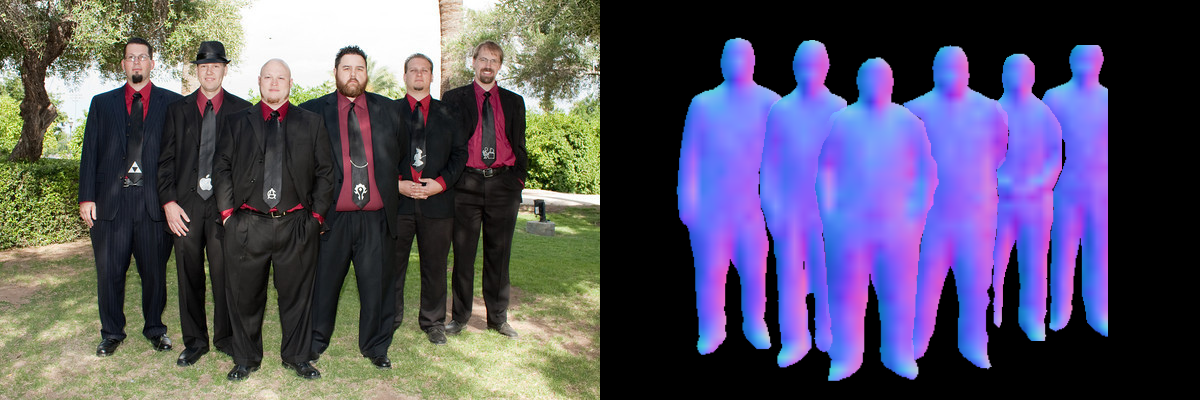}}
  \centering
  \adjustbox{height=\myheightDD}
      {\includegraphics[trim={0cm 0cm 0cm 0cm},clip]{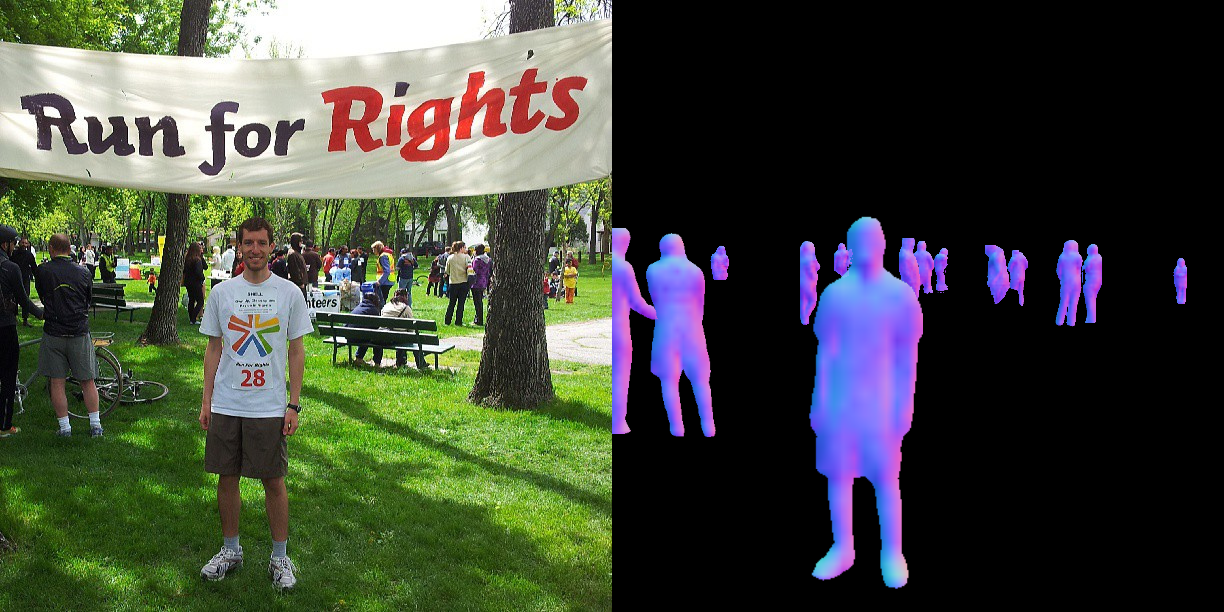}}      
  \adjustbox{height=\myheightDD}
      {\includegraphics[trim={0cm 0cm 0cm 0cm},clip]{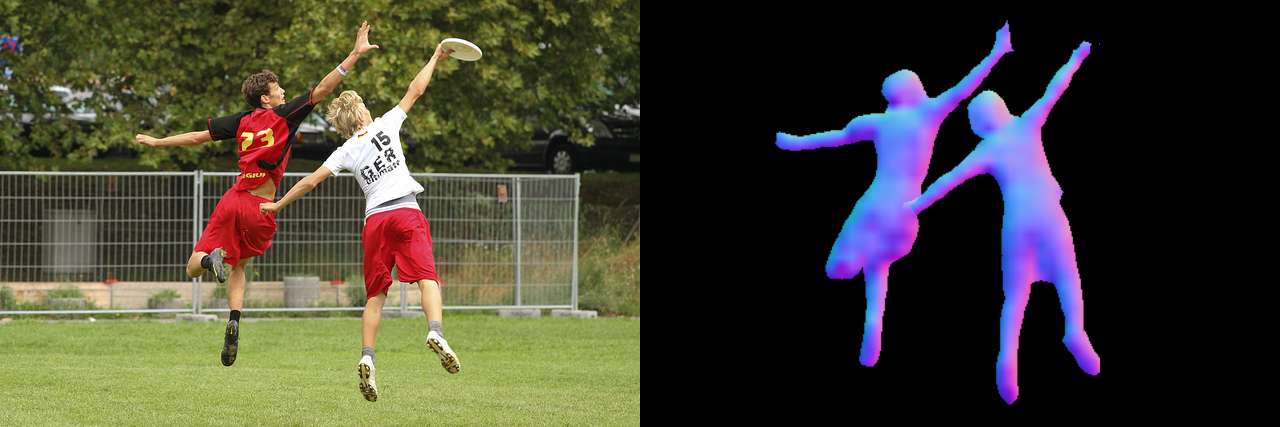}}
  \adjustbox{height=\myheightDD}
      {\includegraphics[trim={0cm 0cm 0cm 0cm},clip]{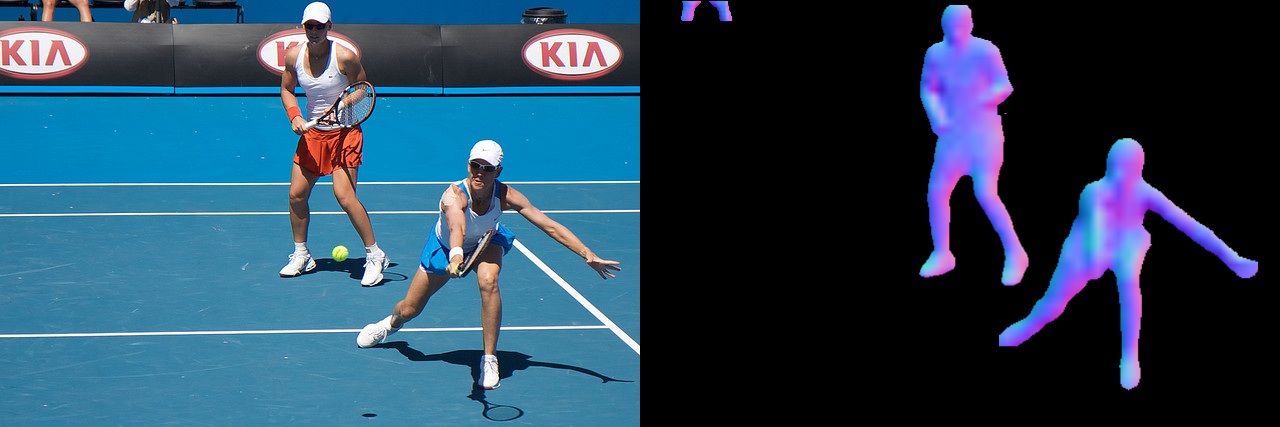}}
  \centering
  \adjustbox{height=\myheightEE}
      {\includegraphics[trim={0cm 0cm 0cm 0cm},clip]{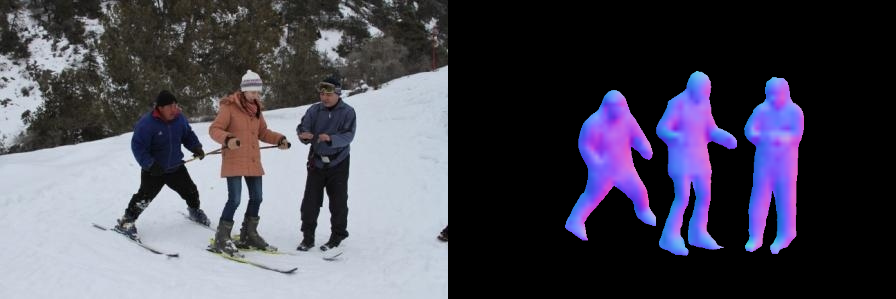}}    
  \adjustbox{height=\myheightEE}
      {\includegraphics[trim={0cm 0cm 0cm 0cm},clip]{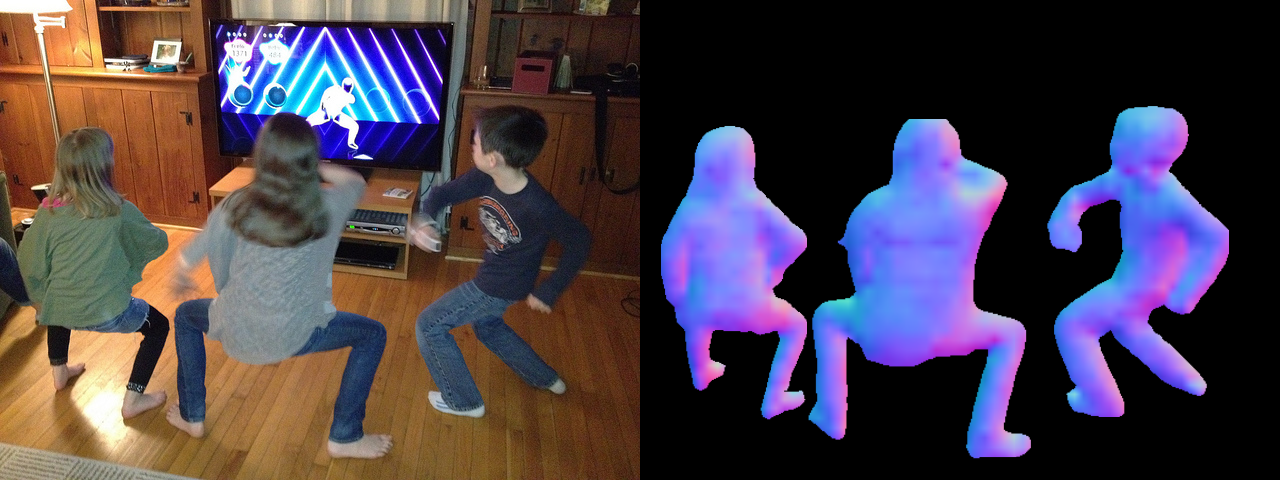}}
  \adjustbox{height=\myheightEE}
      {\includegraphics[trim={0cm 0cm 0cm 0cm},clip]{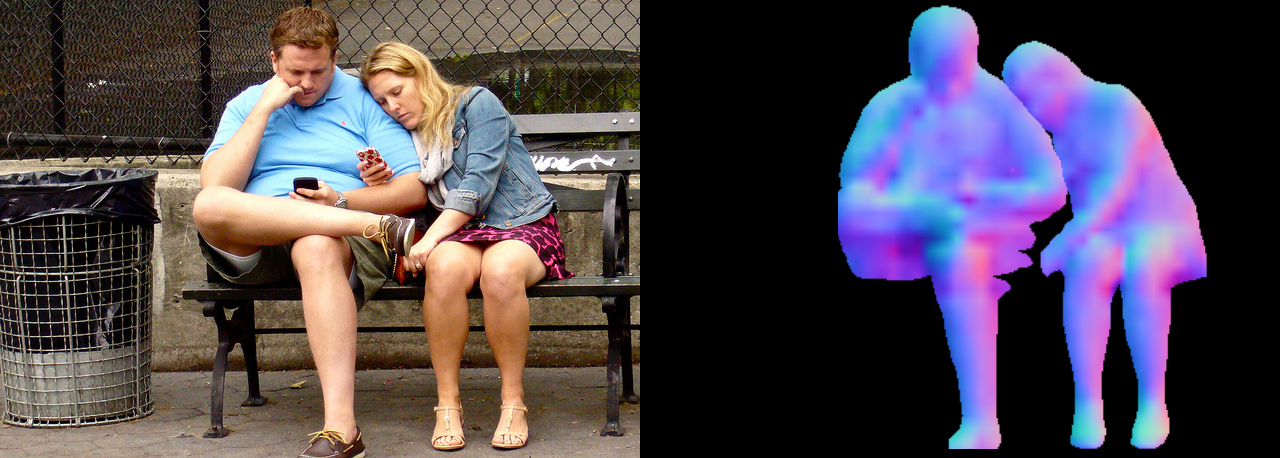}}
  \centering
  \adjustbox{height=\myheightGG}
      {\includegraphics[trim={0cm 0cm 0cm 0cm},clip]{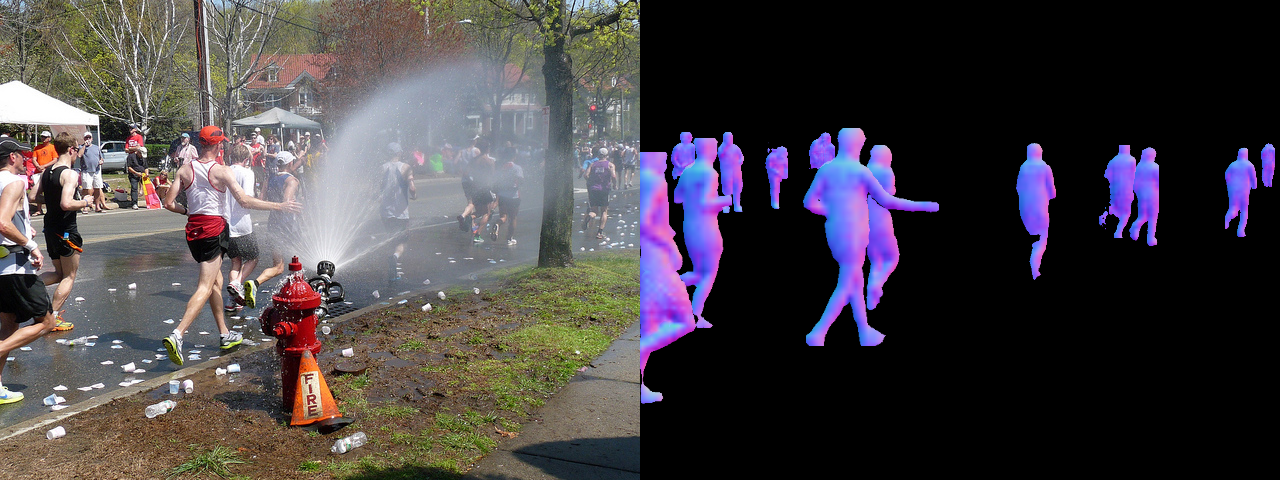}}    
  \adjustbox{height=\myheightGG}
      {\includegraphics[trim={0cm 0cm 0cm 0cm},clip]{figures_normal/mask_538319.png}}
  \caption{SimPose's 3D surface normal predictions on MSCOCO dataset.}
  \label{fig:results_normal_array}
  
  \centering

    \begin{subfigure}{.16666\textwidth}
    \includegraphics[width=1\linewidth,trim={0cm 0cm 0.0cm 0cm},clip]{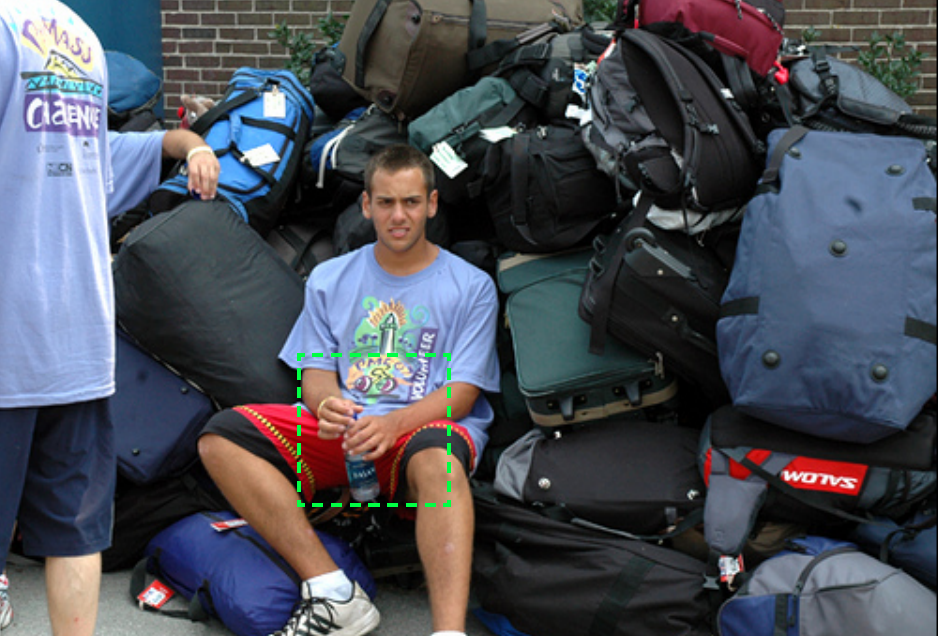}
    \caption*{}
    \end{subfigure}%
    \begin{subfigure}{.16666\textwidth}
    \includegraphics[width=1\linewidth,trim={0cm 0cm 0.0cm 0cm},clip]{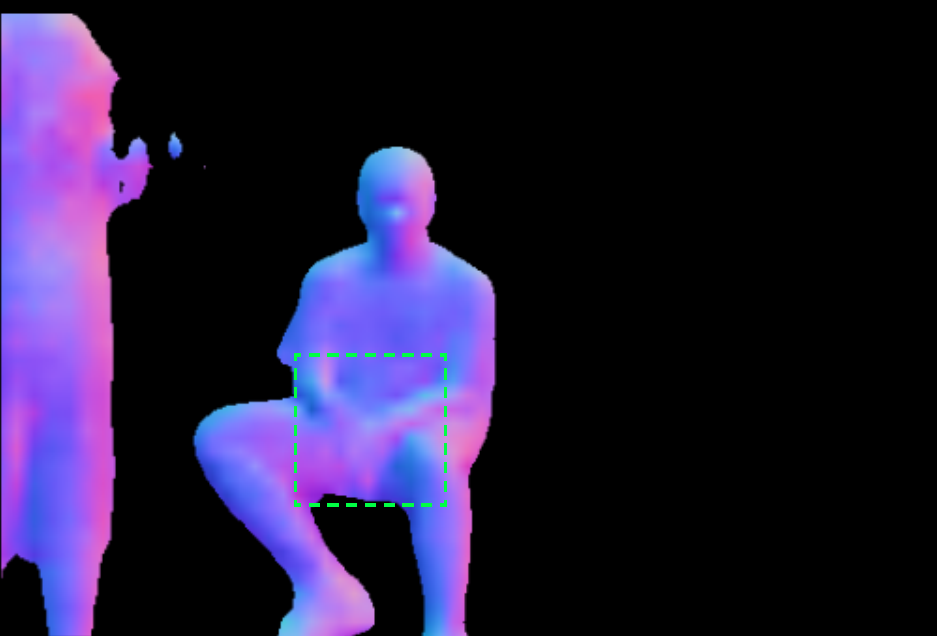}
    \caption*{SMPL}
    \end{subfigure}%
    \begin{subfigure}{.16666\textwidth}
    \includegraphics[width=1\linewidth,trim={0cm 0cm 0.0cm 0cm},clip]{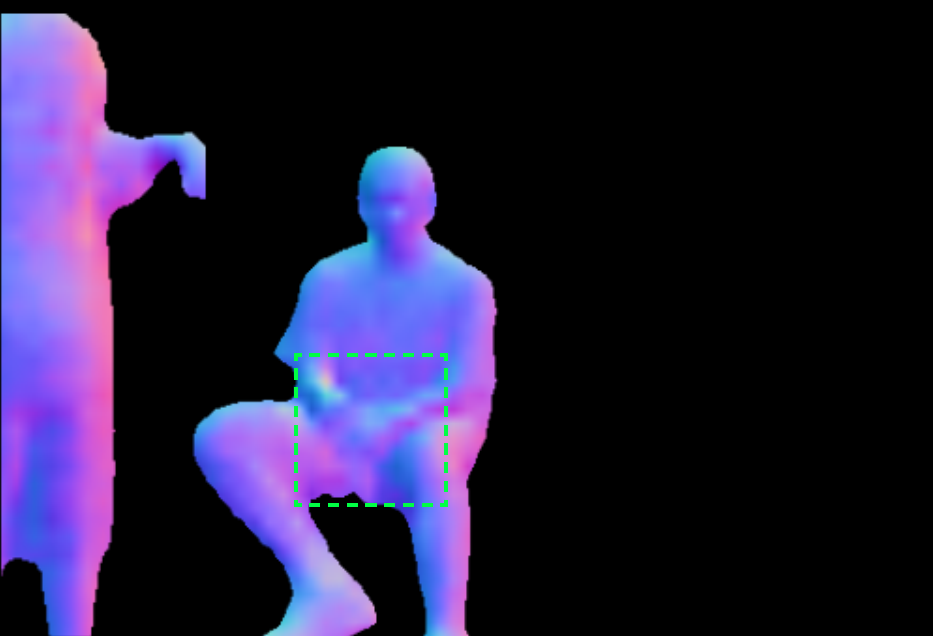}
    \caption*{Renderpeople}
    \end{subfigure}%
    \begin{subfigure}{.16666\textwidth}
    \includegraphics[width=1\linewidth,trim={0cm 0cm 0.0cm 0cm},clip]{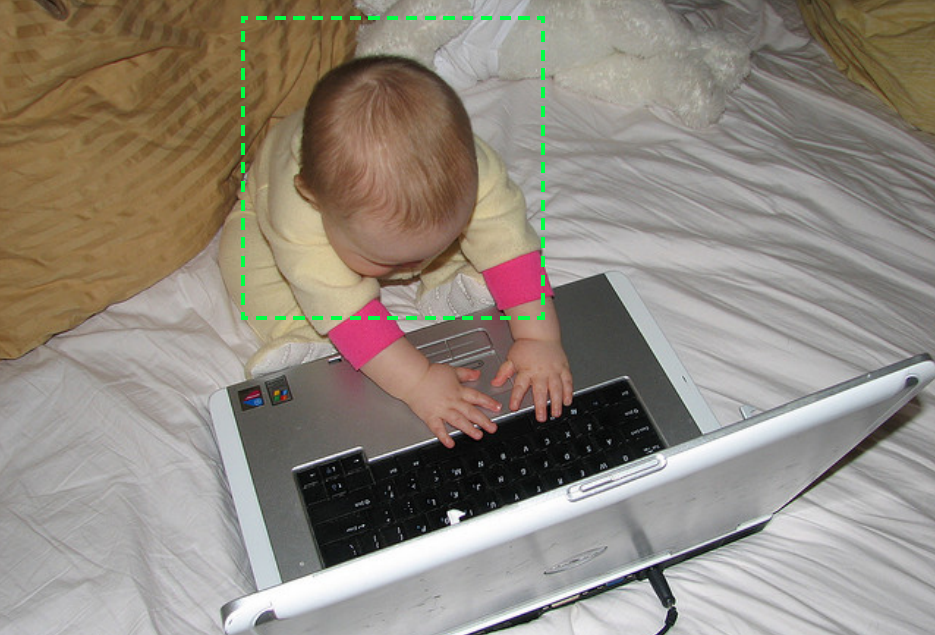}
    \caption*{}
    \end{subfigure}%
    \begin{subfigure}{.16666\textwidth}
    \includegraphics[width=1\linewidth,trim={0cm 0cm 0.0cm 0cm},clip]{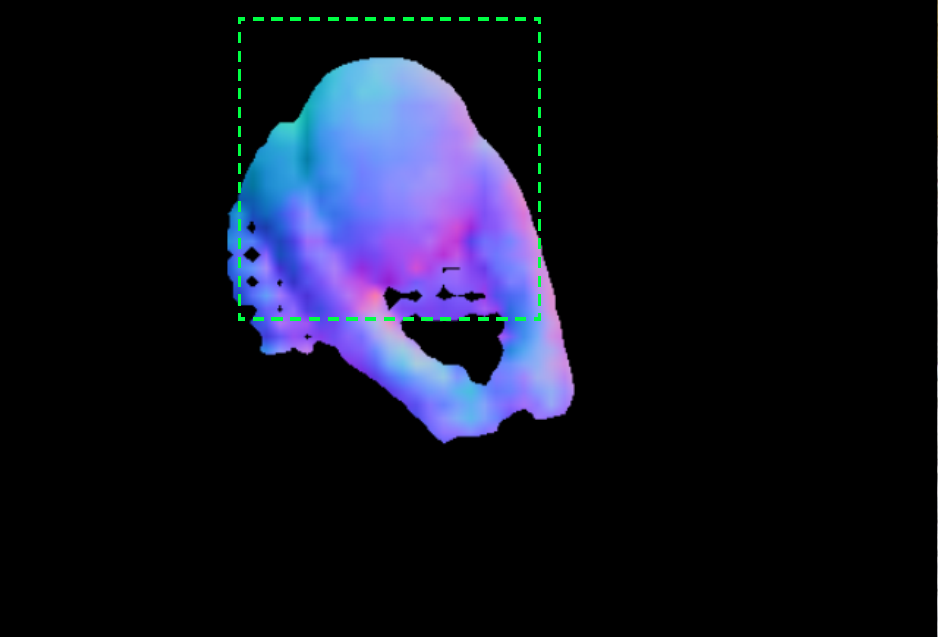}
    \caption*{SMPL}
    \end{subfigure}%
    \begin{subfigure}{.16666\textwidth}
    \includegraphics[width=1\linewidth,trim={0cm 0cm 0.0cm 0cm},clip]{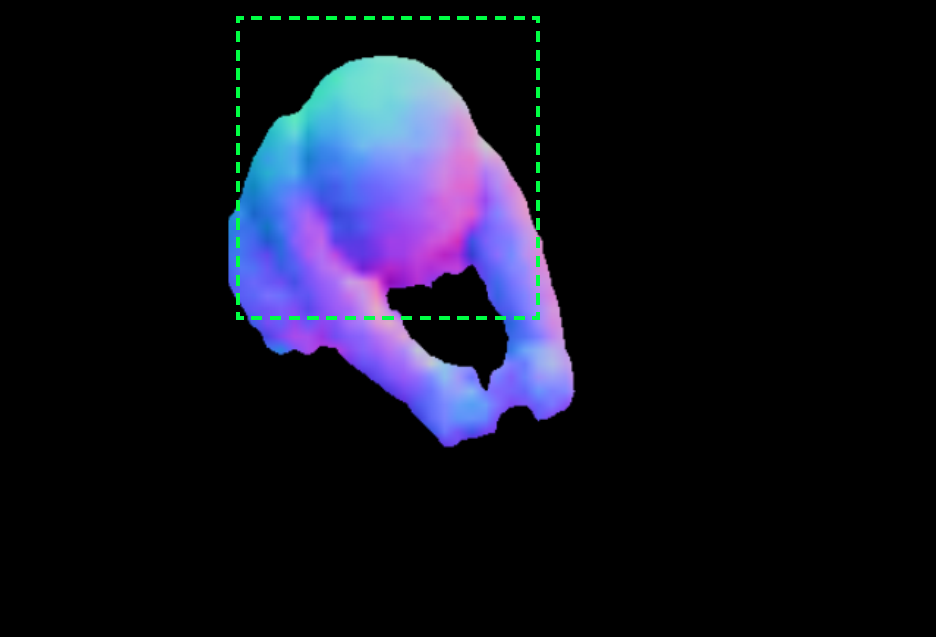}
    \caption*{Renderpeople}
    \end{subfigure}%

  \caption{Comparison of normal predictions trained from different 3D model sources.}
  \label{fig:results_normal_array_side_by_side}
\end{figure*}

%% file: conclusion.tex
\section{Conclusion}
We have shown that our SimPose approach achieves more accurate in-the-wild multi-person dense pose prediction without using any real DensePose labels for training. It also learns 3D human surface normal estimation from only simulated labels and can predict 3D human surface normal on the in-the-wild MSCOCO dataset that currently lacks any surface normal labels. Our SimPose approach achieves this using only 17 non-statistical 3D human figures. We hope the creation process of our simulated dataset and the proposed training scheme opens doors for training other accurate 2.5D and 3D human pose and shape estimation models without manually collecting real world annotations, and still generalizes for accurate multi-person prediction in-the-wild usage. \\

\noindent\textbf{Acknowledgements}
We would like to thank Nori Kanazawa for reviewing our implementation, Fabian Pedregosa and Avneesh Sud for proofreading our paper.